%% file: main.tex
\documentclass{article}
\pdfoutput=1  

    \PassOptionsToPackage{numbers, compress}{natbib}

\usepackage[preprint]{neurips_2023}  

\usepackage[utf8]{inputenc} 
\usepackage[T1]{fontenc}    
\usepackage{hyperref}       
\usepackage{url}            
\usepackage{booktabs}       
\usepackage{amsfonts}       
\usepackage{nicefrac}       
\usepackage{microtype}      
\usepackage{xcolor}         

\usepackage{bm}
\usepackage{amsmath,amssymb,mathtools}
\usepackage{graphicx} 
\usepackage{caption} 
\usepackage[subrefformat=parens]{subcaption}
\usepackage{comment}
\usepackage{listings}

\input{math_commands}

\input{symbols}

\RequirePackage{algorithm}
\RequirePackage{algorithmic}

\title{iMixer: hierarchical Hopfield network implies \\ an invertible, implicit and iterative MLP-Mixer}
\author{%
    Toshihiro Ota  \\
    CyberAgent  \\
    \texttt{ota\_toshihiro@cyberagent.co.jp}  \\
    \And
    Masato Taki  \\
    Rikkyo University  \\
    \texttt{taki\_m@rikkyo.ac.jp} \\
}

\begin{document}

\maketitle

\input{abstract}
\input{sec1}
\input{sec2}
\input{sec3}
\input{sec4}
\input{sec5}
\input{sec6}

\bibliography{references}
\bibliographystyle{nips}

\newpage
\appendix
\input{appendix}

\end{document}

%% file: math_commands.tex
\usepackage{amsmath,amsfonts,bm}

\def\eqref#1{equation~\ref{#1}}

\def\1{\bm{1}}

\def\eps{{\epsilon}}

\def\rb{{\textnormal{b}}}

\def\rh{{\textnormal{h}}}

\def\rt{{\textnormal{t}}}

\def\vb{{\bm{b}}}

\def\vh{{\bm{h}}}

\def\vt{{\bm{t}}}

\DeclareMathAlphabet{\mathsfit}{\encodingdefault}{\sfdefault}{m}{sl}
\SetMathAlphabet{\mathsfit}{bold}{\encodingdefault}{\sfdefault}{bx}{n}

\newcommand{\R}{\mathbb{R}}

\let\ab\allowbreak

%% file: symbols.tex
\newcount\K 
\def\blah{
    \K=0 \loop\ifnum\K<20 
    {\color[rgb]{0.8, 0.8, 0.8} blah blah blah blah blah blah blah blah blah blah} 
    \advance\K by1\repeat 
}

\usepackage{mleftright}
\mleftright

\newcommand{\kh}{generalized Hopfield network}
\newcommand{\kro}{hierarchical Hopfield network}

\newcommand{\xint}{x}

\newcommand{\xii}[2]{\xi^{(#1, #2)}}

\DeclareMathOperator{\LN}{\mathrm{LayerNorm}}
\DeclareMathOperator{\gelu}{\mathrm{GELU}}
\DeclareMathOperator{\fpa}{\mathrm{FPA}}

\newcommand{\imodel}{iMixer}
\newcommand{\iblock}{iMLP}

\newcommand{\hratio}{h_r}  
\newcommand{\niter}{n}  

\newcommand{\eref}[1]{Eq.~(\ref{#1})}
\newcommand{\fref}[1]{Fig.~\ref{#1}}
\newcommand{\tref}[1]{Table~\ref{#1}}
\newcommand{\sref}[1]{Sec.~\ref{#1}}
\newcommand{\aref}[1]{Appendix~\ref{#1}}

\newcommand{\del}{\partial}

\newcommand{\dd}[2]{\frac{d #1}{d #2}}
\newcommand{\dpdp}[2]{\frac{\partial #1}{\partial #2}}

%% file: abstract.tex
\begin{abstract}

In the last few years, the success of Transformers in computer vision has stimulated the discovery of many alternative models that compete with Transformers, such as the MLP-Mixer.
Despite their weak inductive bias, these models have achieved performance comparable to well-studied convolutional neural networks.
Recent studies on modern Hopfield networks suggest the correspondence between certain energy-based associative memory models  
and Transformers or MLP-Mixer, and shed some light on the theoretical background of the Transformer-type architectures design.
In this paper, we generalize the correspondence to the recently introduced \kro, and find \emph{\imodel}, a novel generalization of MLP-Mixer model.
Unlike ordinary feedforward neural networks, \imodel ~involves MLP layers that propagate forward from the output side to the input side.
We characterize the module as an example of invertible, implicit, and iterative mixing module.
We evaluate the model performance with various datasets on image classification tasks, and find that \imodel,
despite its unique architecture, exhibits stable learning capabilities and achieves performance comparable to or better than the baseline vanilla MLP-Mixer.
The results imply that the correspondence between the Hopfield networks and the Mixer models serves as a principle for understanding a broader class of Transformer-like architecture designs.

\end{abstract}

%% file: sec1.tex
\section{Introduction}

\begin{figure}[t]
    \begin{minipage}[b]{0.55\linewidth}
        \centering
        \includegraphics[keepaspectratio, scale=0.25]{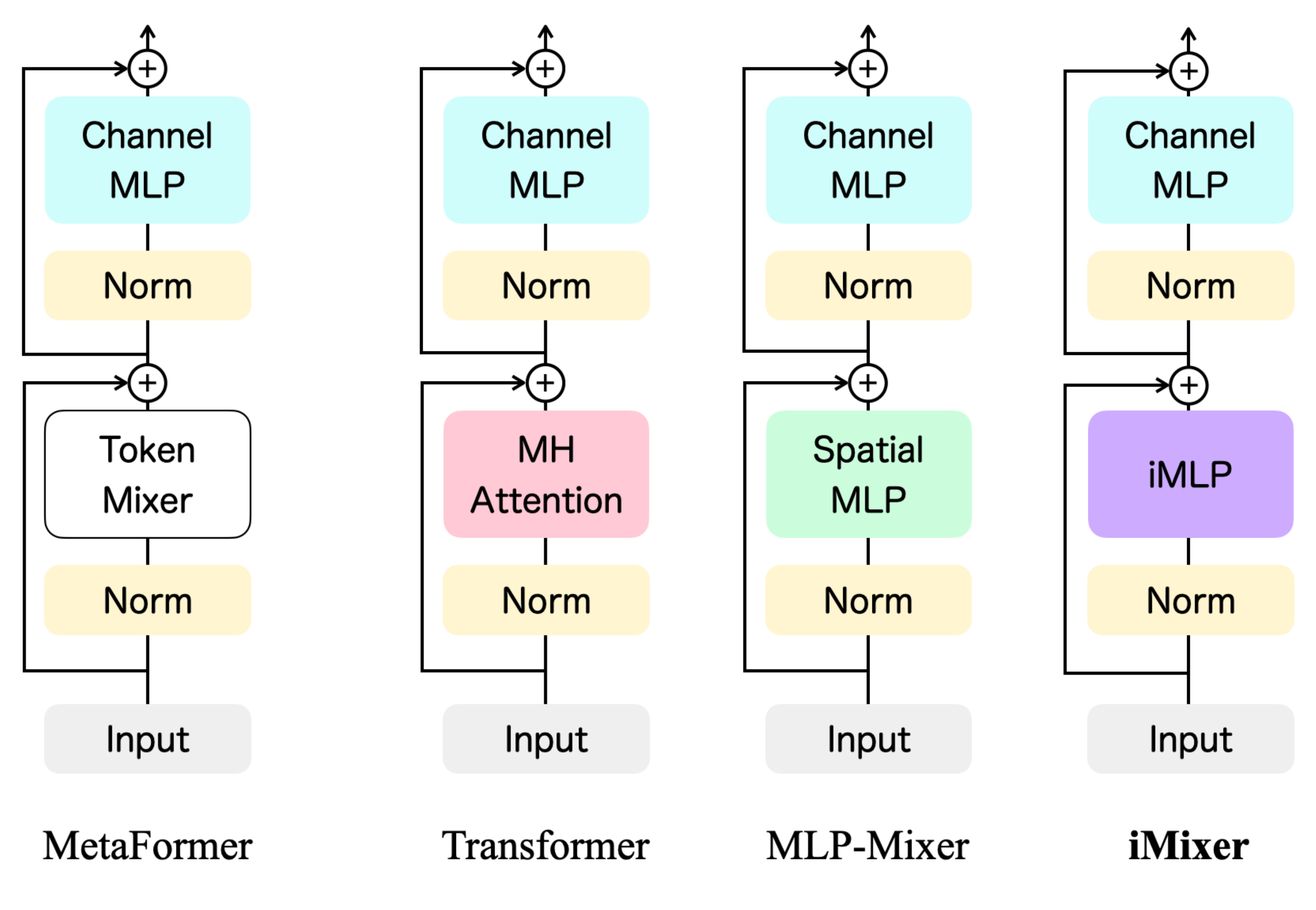}
        \subcaption{MetaFormers}
        \label{fig:metaformers}
    \end{minipage}
    \begin{minipage}[b]{0.22\linewidth}
        \centering
        \includegraphics[keepaspectratio, scale=0.3]{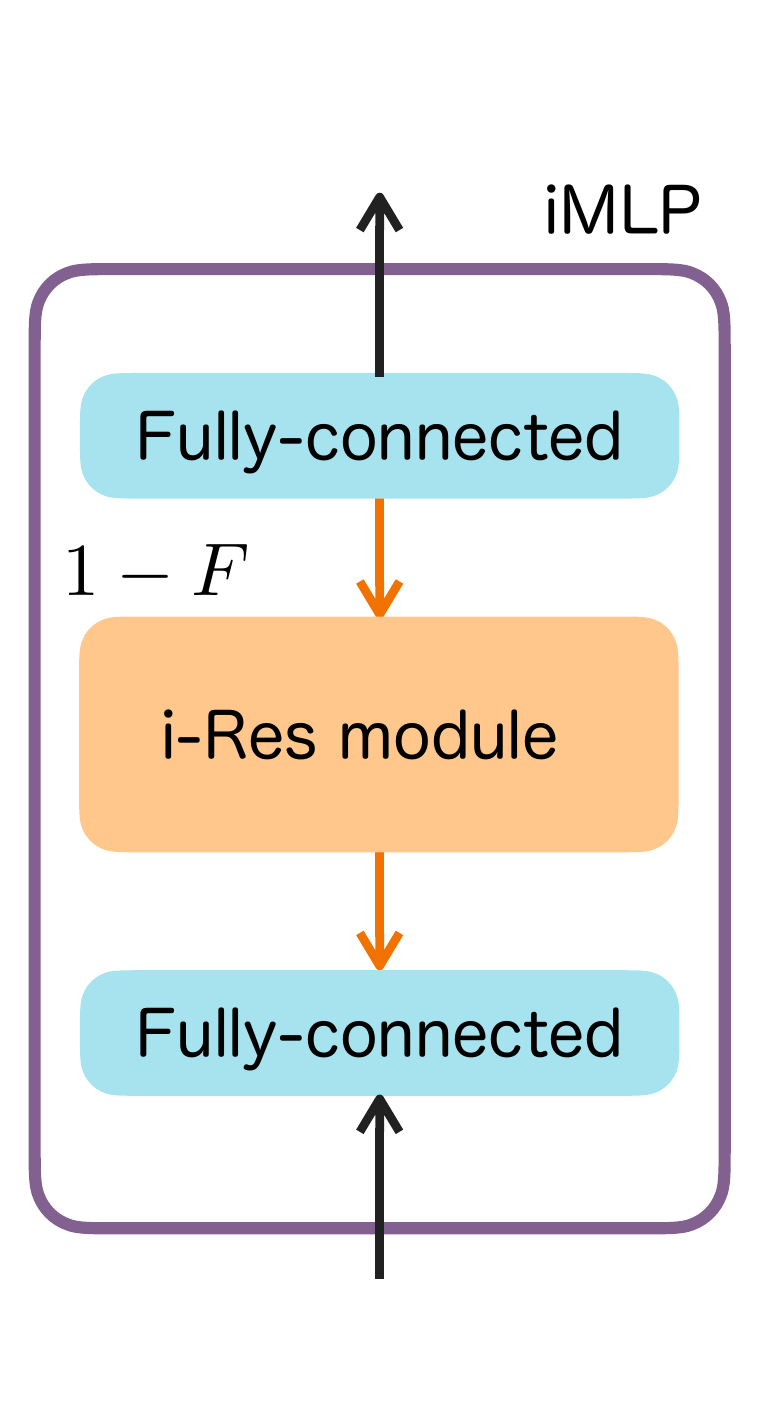}
        \subcaption{\iblock ~module}
        \label{fig:imlp}
    \end{minipage}
    \begin{minipage}[b]{0.22\linewidth}
        \centering
        \includegraphics[keepaspectratio, scale=0.3]{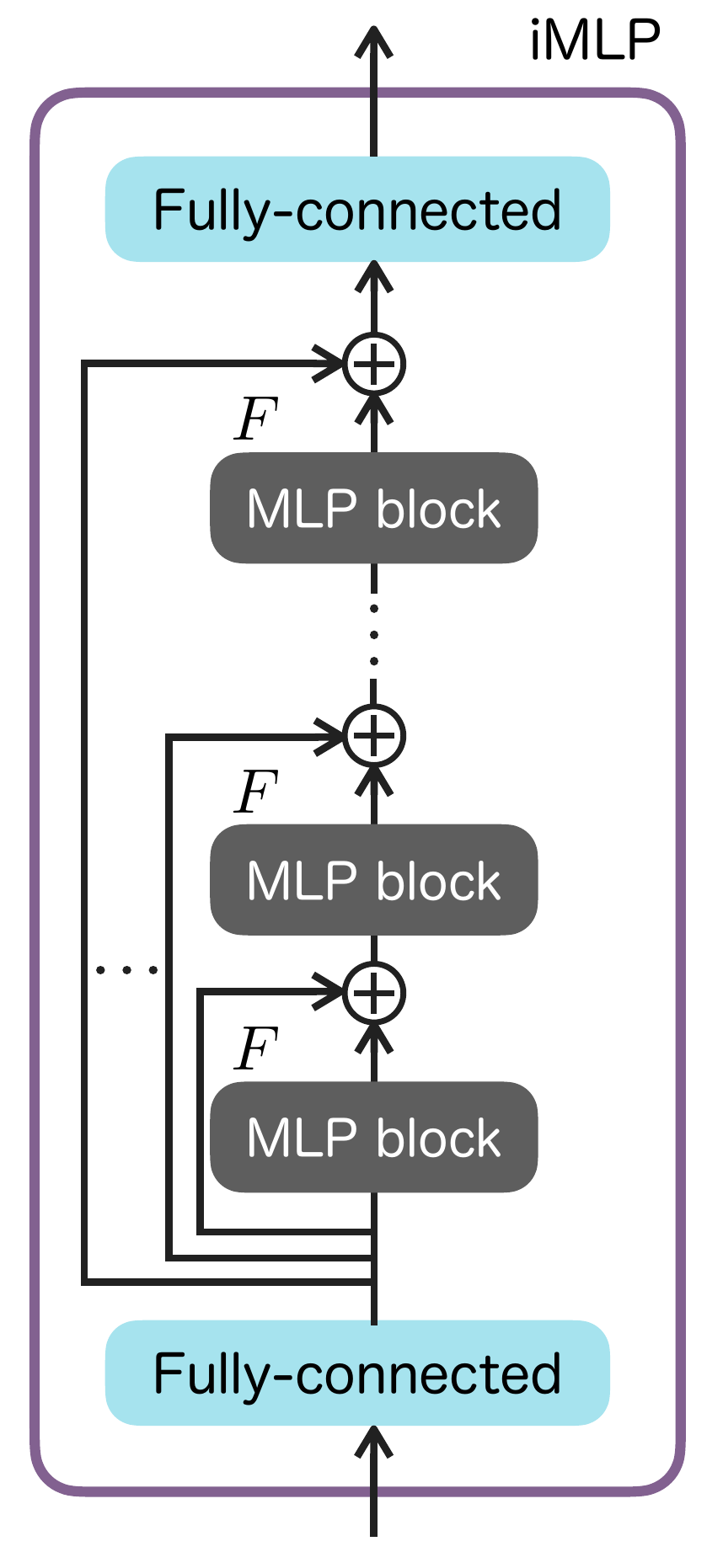}
        \subcaption{$\fpa$}
        \label{fig:fig_iterativemlps}
    \end{minipage}
    \caption{(a) Macro-architecture design of the Transformer-type networks and (b) the structure of \iblock ~module.
            (c) Restricting to invertible $(1-F)$ yields the feedforward realization of the \iblock ~module using the fixed-point approximation ($\fpa$).}
\end{figure}

The Transformer architecture \cite{NIPS2017_3f5ee243}, which has achieved great success in natural language processing in recent years, has proven to perform well in a wide range of computer vision tasks such as image recognition and object detection.
One example is the Vision Transformer (ViT) \cite{DosovitskiyICLR2021,TouvronICML2021}, a standard Transformer architecture applied to image classification that achieves performance competitive with state-of-the-art convolutional neural networks (CNNs).
More recently, alternative models of ViT have been discovered that do not use attention mechanism such as the MLP-Mixer \cite{tolstikhin2021mlp,melas2021you}.
MLP-Mixer learns long-range dependencies between tokens (patches) using only a simple MLP instead of attention mechanism.
Such alternative models are now collectively referred to as MetaFormers \cite{yu2022metaformer,yu2022metaformer2}, see \fref{fig:metaformers}.
Although a number of ViT variants have been proposed and the performance of them continues to improve, developments in their architectural design are somewhat heuristic and a prospective perspective for design is desired.

The Hopfield network is a classical associative memory model of neural networks \cite{hopfield82,hopfield84}.
Recent studies on modern Hopfield networks suggest the correspondence between certain energy-based associative memory models  
and Transformers or MLP-Mixer \cite{krotov2020large,ramsauer2021hopfield,tang2021remark}, which is a promising candidate for a unified description of Transformer's architectural design.
In this paper, we extend the correspondence to the recently introduced \kro ~\cite{krotov2021hierarchical} and show that this extension results in a new class of MLP-Mixer architectures, the \emph{\imodel}.
The macro-architecture design of \imodel ~is the same as that of MLP-Mixer and Transformers (MetaFormers), consisting of stacked token- and channel-mixing blocks.
However, \imodel ~differs significantly from MLP-Mixer and ViT in that the token-mixing block is composed of the invertible ResNet (i-Res) module, or infinitely deep neural network \cite{behrmann2019invertible,bai2019deep,ghaoui2021implicit}, see \fref{fig:imlp}.
Although this architectural design is completely unexpected from the computer vision perspective, it is a natural consequence of multi-layering from the Hopfield network side.
By exploiting the correspondence, we systematically obtain models that have been missed in previous architectural designs.

We show that the dynamics of the \kro ~implies \imodel ~whose token-mixing module is defined as an MLP $f(\cdot)$ that propagates forward from the output side $y$ to the input side $x$ as $x=f(y)$.
The module that cannot be explicitly written down in a forward propagation equation from the input side to the output side is an example of the implicit neural network \cite{ghaoui2021implicit}.
Our \imodel ~is a new type of MetaFormer that uses a mixing module composed of implicit neural networks.
In our case, the implicit structure can be written by inserting $x=f(y)$, a forward propagation in the opposite direction.
Using the i-Res module, this backward-forward propagation can be represented as a normal forward propagating neural network with infinitely iterated layers as shown in \fref{fig:fig_iterativemlps}.
In turn, this representation enables us to easily implement and train the model as a usual feedforward neural network.

Our proposed architecture is not only novel in design, but is actually based on theoretical derivations from the \kro.
By conducting multiple rounds of training on image recognition tasks and obtaining statistics on validation performance,
we also verify that changing the token-mixing modules in MLP-Mixer to the inverted mixing module (\iblock ~module) can improve the performance of the model.
This result suggests a new direction for image recognition architecture design using implicit modules.
The main contributions of the paper is summarized as follows:
\begin{itemize}
    \item The primary contribution is the proposal of a new direction for MetaFormer model design, facilitated by the novel Hopfield/Mixer correspondence. Our intention is to provide a first step towards a theoretical underpinning for MetaFormers.
    \item Our second contribution is the theoretical derivation of a specific new MetaFormer model (\imodel), based on the Hopfield/Mixer correspondence. This work is the first example to \emph{predict} a novel MetaFormer model from modern Hopfield networks.
        In the theoretical formulation, the proposed model naturally incorporates an implicit module $(1-F)^{-1}$, which may initially appear unconventional from a computer vision perspective.
    \item While a theoretical derivation of a deep learning model may be apt to a mere and impractical proposition, our empirical study supports the validity of the proposed \imodel ~as a novel and meaningful MetaFormer, which can be considered our third contribution.
\end{itemize}

%% file: sec2.tex
\section{Related Work}

\noindent \textbf{Transformer \& Mixer models.}
Transformer \cite{NIPS2017_3f5ee243}, which has proven its effectiveness in natural language processing, is now widely used in computer vision.
ViT \cite{DosovitskiyICLR2021} is an alternative model to CNN that uses Transformer encoders for image recognition and demonstrates the effectiveness of Transformer in computer vision.
Various improved versions of ViT have also been proposed, such as incorporating local structure in attention or using distillation \cite{TouvronICML2021}.

While it is widely believed that an attention mechanism is critical to the success of ViT, there are results that cast doubt on its necessity;
MLP-Mixer \cite{tolstikhin2021mlp, touvron2022resmlp, melas2021you, liu2022we} has shown that by simply replacing the attention mechanism of ViT with MLP,
it is possible to achieve performance approaching that of ViT. This discovery has stimulated a series of studies that have shown that a wide range of token-mixing mechanisms,
including pooling \cite{yu2022metaformer}, global filtering \cite{rao2021global}, recurrent layers \cite{tatsunami2022sequencer},
and graph neural networks \cite{han2022vision}, can be used as substitutes for the attention mechanism.
These findings suggest that, in fact, in computer vision, it is not the attention mechanism itself that is important, but the macro-architecture design of the Transformer, which repeatedly mixes tokens and channels.
These groups of models are referred to as MetaFormers \cite{yu2022metaformer, yu2022metaformer2}.

\noindent \textbf{Classical \& modern Hopfield networks.}
The Hopfield network is a well-known model for associative memory in neural networks \cite{hopfield82,hopfield84}.
In this network, stored memories and their retrieval are described by the attractors of an energy function and the dynamics converging to them.
In practice, however, the classical Hopfield network is known to suffer from a limited memory capacity.
To address this issue, models with substantially higher memory capacities have recently been proposed \cite{NIPS2016_eaae339c,demircigil2017model,krotov2018dense},
but they are supposed to have many-body interactions among neurons, which is biologically implausible in the brain.

Close to the time when these modern Hopfield networks were proposed, Ramsauer et al.~found that each of the attention modules in the Transformer
can essentially be identified with the process of the update rule of a certain continuous Hopfield network \cite{ramsauer2021hopfield},
and their algorithm has been developed and applied to some tasks with success \cite{NEURIPS2020_da4902cb,yang2022transformers,hoover2023energy}.
Based on these heuristic findings and earlier works, Krotov and Hopfield developed more comprehensive associative memory models,
which consist of visible and hidden neurons with only two-body interactions between them \cite{krotov2020large,krotov2021hierarchical}.
In their papers, Krotov and Hopfield demonstrated that many of modern neural network models in the literature
can be derived from Hopfield-type networks equipped with Lagrangians that define the systems.

\noindent \textbf{Implicit deep learning.}
Implicit neural networks and implicit layers have been discussed in various papers
\cite{ghaoui2021implicit,bai2019deep,winston2020monotone,bai2021stabilizing,bai2020multiscale,kawaguchi2021theory,pmlr-v162-agarwala22a}.
Implicit layer generally refers to a layer that cannot be written as an explicit forward propagation expression, such as $y=f(x)$, but is expressed as an implicit expression $g(x,y)=0$.

The invertible ResNet layer \cite{behrmann2019invertible} used in this paper is also a layer related to implicit layers.
In this paper, we consider a layer in which the propagation from the output side $y$ to the input side $x$ is represented by the invertible ResNet layer $x=y+F(y)$.
A fixed-point iteration method can be used to represent this layer in the usual orientation.
The resulting forward propagation equation is $y=x-F(x-F(\cdots))$, an infinitely deep network that iteratively adapts the same layer as depicted in \fref{fig:fig_iterativemlps}.
In other words, this forward propagation cannot be expressed in terms of ordinary finite layers.

The deep equilibrium model (DEQ) \cite{bai2019deep, bai2020multiscale} proposed a new computer vision architecture by introducing an implicit layer defined by fixed point for iterative application of a layer.
The iterative layer structure of \imodel ~has similarities with this study.
Unlike DEQ, which uses back propagation with implicit differentiation using the Jacobian, \imodel ~simply trains the entire model end-to-end using the usual gradient descent method.

%% file: sec3.tex
\section{Background}

To fix notations, we here provide a derivation of the MLP-Mixer \cite{tolstikhin2021mlp} from a continuous Hopfield network.

\begin{figure}[t]
    \begin{minipage}[b]{0.33\linewidth}
        \centering
        \includegraphics[keepaspectratio, scale=0.15]{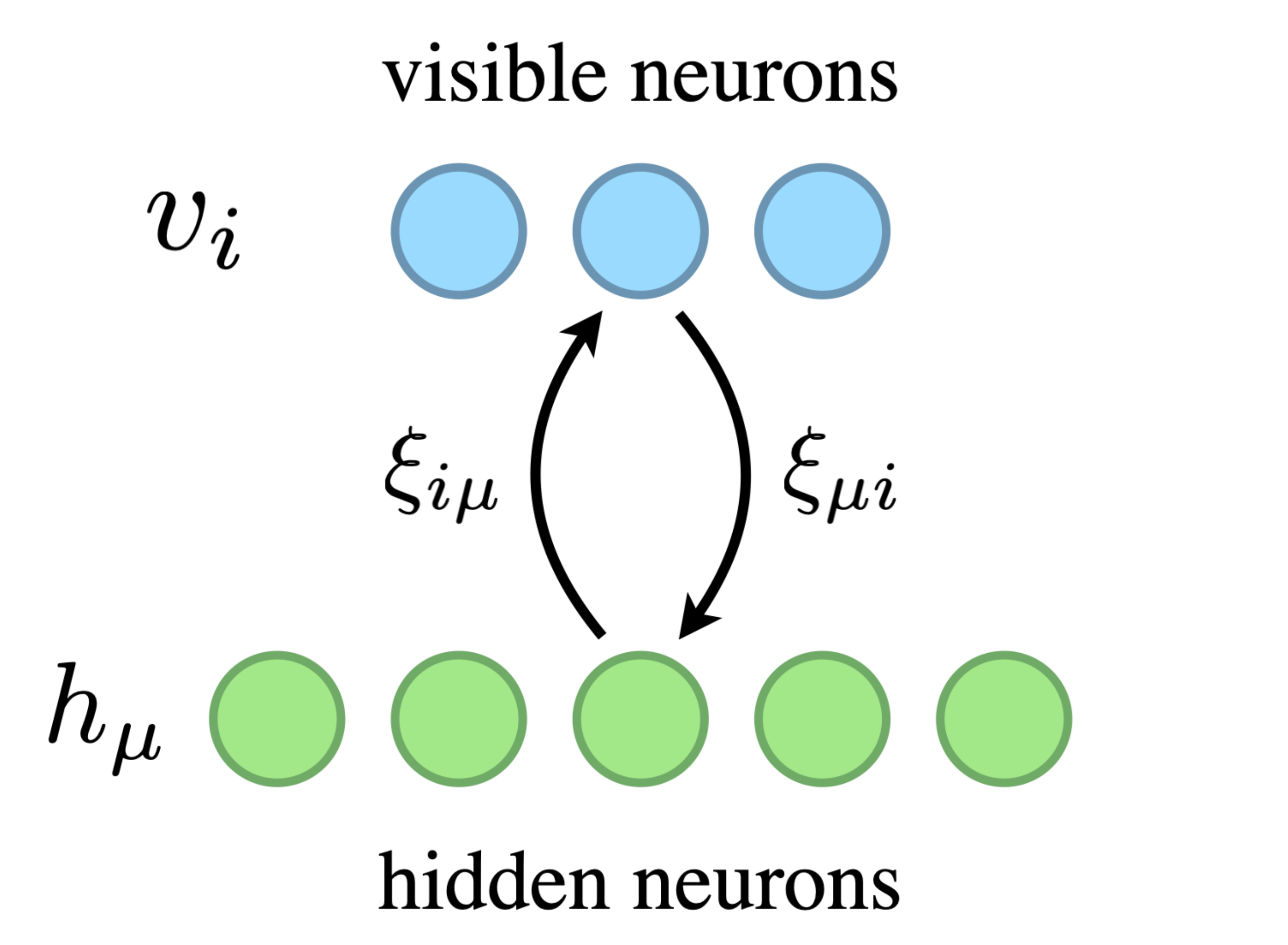}
        \subcaption{Generalized Hopfield network}
        \label{fig:neurons}
    \end{minipage}
    \begin{minipage}[b]{0.33\linewidth}
        \centering
        \includegraphics[keepaspectratio, scale=0.2]{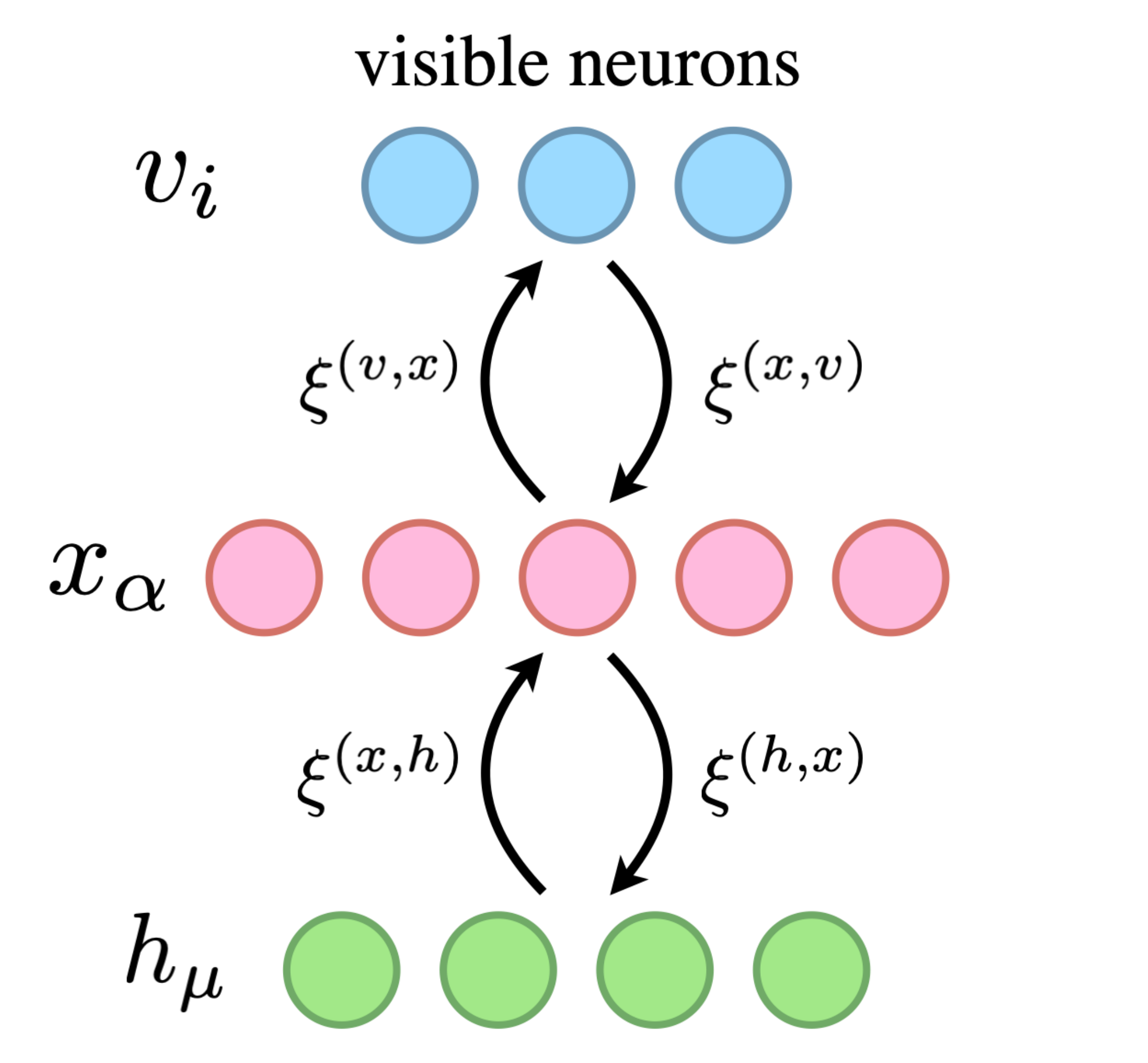}
        \subcaption{Hierarchical Hopfield network}
        \label{fig:hierarchical}
    \end{minipage}
    \begin{minipage}[b]{0.33\linewidth}
        \centering
        \includegraphics[keepaspectratio, scale=0.17]{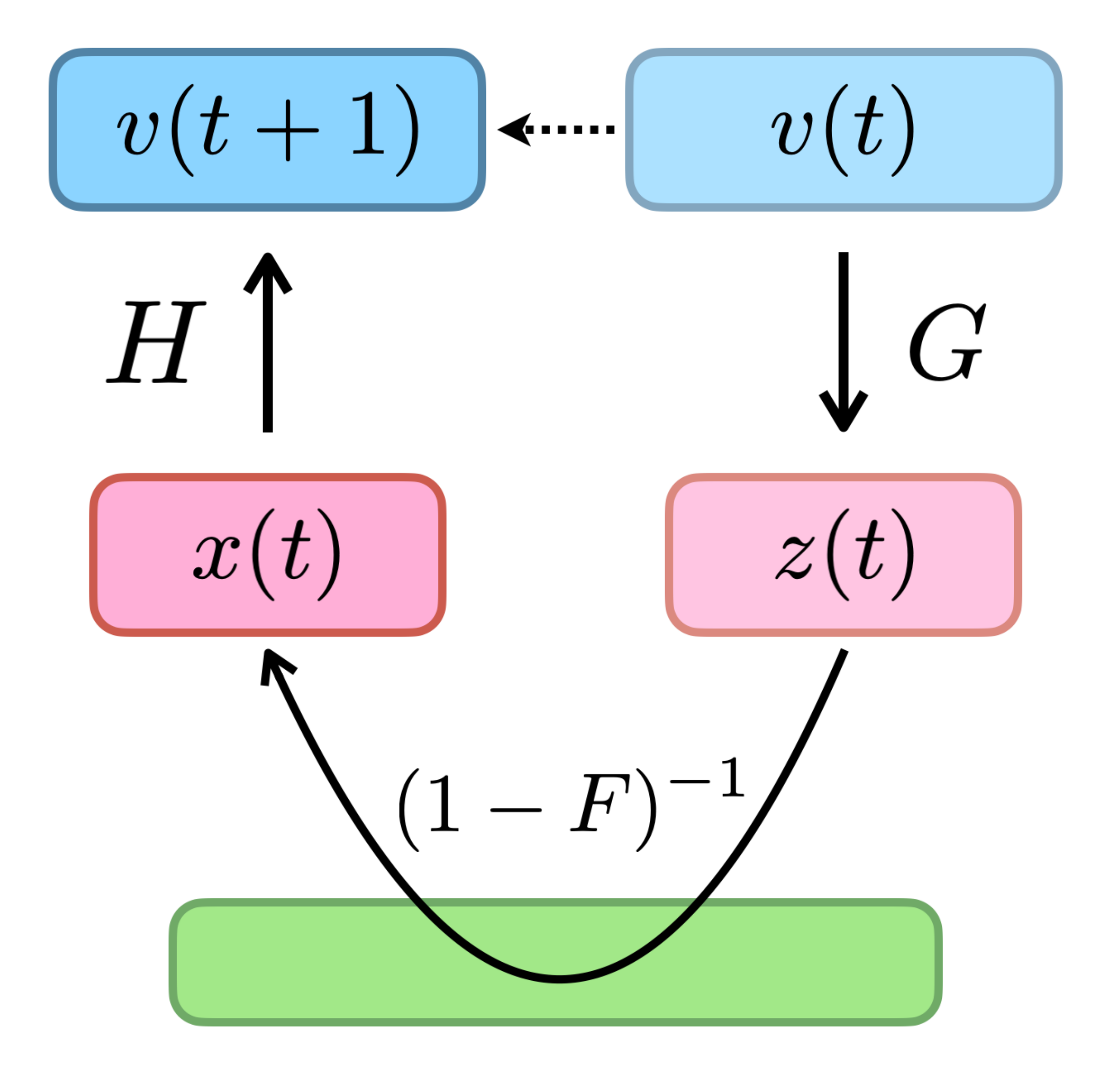}
        \subcaption{Update rule for visible neurons}
        \label{fig:updaterule}
    \end{minipage}
    \caption{Modern Hopfield networks.}
\end{figure}

\subsection{Overview of the \kh}
\label{sec:Overview}

Let us first briefly review the \kh ~proposed in \citep{krotov2020large}.
In this system, the dynamical variables are composed of $N_v$ visible neurons and $N_h$ hidden neurons both continuous,
\begin{equation}
    v(t) \in \R^{N_v},  \quad  h(t) \in \R^{N_h},
\end{equation}
where the argument $t$ can be thought of as ``time''.
The interaction matrices between them,
\begin{equation}
    \xi \in \R^{N_h \times N_v},
    \quad
    \tilde{\xi} \in \R^{N_v \times N_h},
\end{equation}
are basically supposed to be symmetric: $\tilde{\xi} = \xi^{\top}$, see \fref{fig:neurons}.
With the relaxing time constants of the two groups of neurons $\tau_v$ and $\tau_h$, the system is described by the following differential equations,
\begin{align}
    \tau_v \dd{v_{i}(t)}{t}
        &=\sum_{\mu=1}^{N_{h}} \xi_{i \mu} f_{\mu}(h(t)) - v_{i}(t),  \\
    \tau_h \dd{h_\mu(t)}{t}
        &=\sum_{i=1}^{N_{v}} \xi_{\mu i} g_{i}(v(t)) - h_{\mu}(t),  \label{eq:heq}
\end{align}
where the activation functions $f$ and $g$ are determined through Lagrangians
$L_h: \R^{N_h}\to \R$ and $L_v: \R^{N_v}\to \R$, such that
\begin{equation}
    f_\mu(h) = \dpdp{L_{h}(h)}{h_{\mu}},
    \quad
    g_i(v)=\dpdp{L_{v}(v)}{v_{i}}.
\end{equation}

The canonical energy function for this system is given by
\begin{equation}
    E(v, h)
        = \sum_{i=1}^{N_v} v_{i} g_{i}(v) - L_{v}(v)  
        + \sum_{\mu=1}^{N_h} h_{\mu} f_{\mu}(h) - L_{h}(h)
        - \sum_{\mu, i} f_{\mu} \xi_{\mu i} g_{i}.
\end{equation}
One can easily find that this energy function monotonically decreases along the trajectory of the dynamical equations to define an associative memory model,
\begin{equation}
    \dd{E(v(t), h(t))}{t} \leq 0,
\end{equation}
provided that the Hessians of the Lagrangians are positive semi-definite.
In addition to this, if the overall energy function is bounded from below, the trajectory is guaranteed to converge to a fixed point attractor state, which corresponds to one of the local minima of the energy function.
Such fixed points and the process of convergence are thought of as associative memories and memory retrieval of an associative memory model.
The formulation of neural networks in terms of Lagrangians and the associated energy functions enables us to easily experiment with different choices of the activation functions and different architectural arrangements of neurons.

\subsection{MLP-Mixer as an associative memory model}
\label{sec:Mixer}

Suppose we have a fixed interaction matrix $\xi_{\mu i}$, then the system is defined by the choice of Lagrangians $L_h$ and $L_v$.
Tang and Kopp demonstrated that the specific choice of Lagrangians called ``model C'' in \citep{krotov2020large}
essentially reproduces the mixing layers in the MLP-Mixer \citep{tang2021remark}, which is given by the following Lagrangians:
\begin{equation}
    L_h(h) = \sum_{\mu} \phi(h_\mu),
    \quad
    L_v(v) = \sqrt{ \sum_i (v_i - \bar{v})^2 },
    \label{eq:Lmixer}
\end{equation}
where $\phi$ will be specified below, and $\bar{v} = \sum_i v_i / N_v$.
For these Lagrangians, the activation functions are
\begin{align}
    f_\mu(h)
        &= \dpdp{L_h}{h_\mu}
        = \phi'(h_\mu),  \\
    g_i(v)
        &= \dpdp{L_v}{v_i}
        = \frac{v_i - \bar{v}}{\sqrt{\sum_j (v_j - \bar{v})^2}}
        = \LN(v)_i.
\end{align}

We now consider the adiabatic limit, $\tau_v \gg \tau_h$, which means that the dynamics of the hidden neurons is much faster than that of the visible neurons,
i.e., we can take $\tau_h \to 0$:
\begin{equation}
    \text{\eref{eq:heq}}
    \quad \rightsquigarrow \quad
    h_{\mu}(t) = \sum_{i=1}^{N_{v}} \xi_{\mu i} \LN(v(t))_i.
\end{equation}
Substituting the above expression into the other dynamical equation, we find
\begin{equation}
    \tau_v \dd{v_{i}(t)}{t}
        =\sum_\mu \xi_{i \mu} \phi'\bigg( \sum_{j} \xi_{\mu j} \LN(v(t))_j \bigg) - \alpha v_{i}(t).
\end{equation}
Notice that we can put an arbitrary coefficient $\alpha$, which can be even zero, in front of the decay term,
since for this choice of Lagrangian $L_v$ its Hessian has a zero mode:
\begin{equation}
    \sum_j M_{ij}(v_j - \bar{v}) = 0,
    \quad
    M_{ij} :=\frac{\del^2 L_v}{\del v_i \del v_j}.
\end{equation}

If we take $\alpha=0$ and discretize the differential equation by taking $\Delta t = \tau_v$, then we obtain the update rule for the visible neurons,
\begin{equation}
    v_i(t+1) = v_i(t) + \sum_\mu \xi_{i \mu} \, \sigma\bigg( \sum_{j} \xi_{\mu j} \LN(v(t))_j \bigg),
\end{equation}
where we defined $\sigma := \phi'$.
If one chooses $\sigma=\gelu$, this update rule is identified with the token- and channel-mixing blocks in the mixing layers discussed in \citep{tolstikhin2021mlp}.
We will utilize this fact in \sref{sec:Derivation}.

%% file: sec4.tex
\section{Model}
\label{sec:model}

Along the line of \cite{krotov2020large}, Krotov further extended the model in such a way that the \kh ~can consist of multiple hidden layers \cite{krotov2021hierarchical}, which we refer to as the \kro.
Based on the result in \sref{sec:Mixer} and Krotov's extension, in this section we propose an invertible, implicit and iterative MLP-Mixer (\imodel), a generalization of the MLP-Mixer with multiple hidden layers.

As one of the simplest generalizations in view of the \kro, we consider the case of three layers in total,%
\footnote{For more general cases, see \aref{appendix:general}.}  
which consists of one visible layer with $N_v$ neurons and two hidden layers with $N_{\xint}$ and $N_h$ neurons for each, as shown in \fref{fig:hierarchical}.
The dynamics of this system is then described by the following differential equations \cite{krotov2021hierarchical},
\begin{align}
    \tau_v \dd{v_{i}(t)}{t}
        &=\sum_{\alpha=1}^{N_{\xint}} \xii{v}{\xint}_{i \alpha} e_{\alpha}(\xint(t)) - v_{i}(t),  \\
    \tau_\xint \dd{\xint_{\alpha}(t)}{t}
        &=\sum_{i=1}^{N_{v}} \xii{\xint}{v}_{\alpha i} g_{i}(v(t)) + \sum_{\mu=1}^{N_{h}} \xii{\xint}{h}_{\alpha \mu} f_{\mu}(h(t))- \xint_{\alpha}(t),  \\
    \tau_h \dd{h_\mu(t)}{t}
        &=\sum_{\alpha=1}^{N_{\xint}} \xii{h}{\xint}_{\mu \alpha} e_{\alpha}(\xint(t)) - h_{\mu}(t),
\end{align}
where $\xii{A}{B}$ indicates the interaction from $B$ neurons to $A$ neurons: $\xii{A}{B}\in \R^{N_A\times N_B}$,
and the activation functions are again determined through each of the Lagrangian,
\begin{align}
    f_\mu(h) = \dpdp{L_{h}(h)}{h_{\mu}},
    \quad
    g_i(v)=\dpdp{L_{v}(v)}{v_{i}},
    \quad
    e_\alpha(x)=\dpdp{L_{x}(x)}{x_{\alpha}}.
\end{align}

\subsection{iMixer: invertible, implicit and iterative MLP-Mixer}
\label{sec:Derivation}

In order to derive \imodel, we take our Lagrangians as follows:
\begin{equation}
    L_h(h) = \sum_{\mu} \phi_h(h_\mu),
    \quad
    L_v(v) = \sqrt{ \sum_i (v_i - \bar{v})^2 },
    \quad
    L_x(x) = \sum_{\alpha} \phi_x(x_\alpha),
\end{equation}
where $\bar{v} = \sum_i v_i / N_v$ and the activation function for the visible neurons reads $g_i=\LN(\cdot)_i$.
Hence by following the same discussion as in \sref{sec:Mixer}, the dynamical equation for the visible neurons eventually becomes a residual network,
\begin{equation}
    v_i(t+1) = v_i(t) + \sum_{\alpha=1}^{N_{\xint}} \xii{v}{\xint}_{i \alpha} e_\alpha(x(t)).  \label{eq:vaffine}
\end{equation}

To obtain the time evolution for $\xint(t)$ and to solve the whole update rule, we again consider the adiabatic limit, $\tau_v \gg \tau_\xint \gg\tau_h$,
then the dynamical equations for the hidden neurons are reduced to
\begin{align}
    \xint_{\alpha}(t)
        &=\sum_{i=1}^{N_{v}} \xii{\xint}{v}_{\alpha i} \LN(v(t))_{i}
            + \sum_{\mu=1}^{N_{h}} \xii{\xint}{h}_{\alpha \mu} f_{\mu}(h(t)),  \\
    h_{\mu}(t)
        &=\sum_{\alpha=1}^{N_{\xint}} \xii{h}{\xint}_{\mu \alpha} e_{\alpha}(\xint(t)).  \label{eq:hred}
\end{align}
By substituting \eref{eq:hred} to the other, one obtains
\begin{equation}
    \xint(t) - \xii{\xint}{h} f(\xii{h}{\xint} e(\xint(t)))
        = \xii{\xint}{v} \LN(v(t)),
\end{equation}
where we switched to the vector and matrix notation for simplicity.
Note that all the products here are matrix multiplication.
For this expression, we define a contractive MLP block $F: \R^{N_\xint} \to \R^{N_\xint}$ and a fully-connected layer $G: \R^{N_v} \to \R^{N_\xint}$ as
\begin{equation}
    F := (\xii{\xint}{h} f) \circ (\xii{h}{\xint} e),
    \quad
    G := \xii{\xint}{v} \LN(\cdot),
\end{equation}
and write
\begin{equation}
    (1 - F)(x(t)) = G(v(t)) =: z(t).  \label{eq:ires}
\end{equation}

Combining Eqs.~(\ref{eq:ires}) and (\ref{eq:vaffine}), we obtain the update rule for the visible neurons depicted as in \fref{fig:updaterule}:
\begin{align}
    v(t+1)   &= v(t) + H(x(t)),  \\
    \xint(t) &= (1 - F)^{-1}(z(t)),  \label{eq:xres}  \\
    z(t)     &= G(v(t)),
\end{align}
where $H := \xii{v}{\xint} e(\cdot)$.
The activation functions $f=\phi_h'$ and $e=\phi_x'$ need not be specified at this stage.
In practice, we will fix them to a neuron-wise activation $f=e=\gelu$ in experiments to compare with the vanilla MLP-Mixer.

To fully solve the update rule for the visible neurons, we need to compute the inverse of the residual connection $(1-F)^{-1}$ in \eref{eq:xres}, which is generically intractable.
To resolve this difficulty, we employ the fixed-point iteration method for residual connections provided by \cite{behrmann2019invertible};
we set $\xint^0 = z(t)$ and perform the fixed-point iterations for some integer $n$ as follows:
\begin{equation}
    \xint^{a+1} = \xint^0 + F(\xint^a),
    \quad
    F = (\xii{\xint}{h} f) \circ (\xii{h}{\xint} e),
    \label{eq:fpamodule}
\end{equation}
for $a = 0,\dots,n-1$.
Behrmann et al.~showed that this sequence of operations makes $\xint^a$ converge to $\xint(t)$ exponentially fast with respect to $n$ owing to the Banach fixed-point theorem.
We thus can approximate $x(t)\simeq x^n$ and numerically solve the update rule, as shown in Algorithm~\ref{alg:imixer}.
Figure~\ref{fig:imlp} illustrates our inverted mixing module (\iblock ~module) using the i-Res module, and the whole \imodel ~architecture is composed of those as token-mixing modules,
\begin{equation}
    v(t+1) = v(t) + \xii{v}{\xint} \gelu \left( \fpa \left( \xii{x}{v} \operatorname{LN}(v(t)) \right) \right),
    \label{eq:imlp}
\end{equation}
where the activation functions $f$ and $e$ are taken to be $\gelu$ here, and $\fpa$ indicates the fixed-point approximation of the operation $(1-F)^{-1}$.
Thus the $\fpa$ layer \eref{eq:fpamodule} in the \iblock ~module that maps $z(t)=\xint^0$ to $x(t)\simeq x^n$
is implemented as a deep feedforward network that applies the same layer $F(\cdot)$ $n$ times as in \fref{fig:fig_iterativemlps}.

\begin{algorithm}[t]
    \caption{Feedforward computation of the \iblock ~module.}
    \label{alg:imixer}
    \begin{algorithmic}
        \STATE {\bfseries Input:} input $x$,
        fully-connected layer $G$, contractive MLP block $F$,
        fully-connected layer $H$, number of fixed-point iterations $n$
        \STATE Init: $x^0 := G(x)$
        \FOR{$a=0, \ldots, n-1$}
        \STATE $x^{a+1} := x^0 + F(x^{a})$
        \ENDFOR
        \STATE {\bfseries return:} $H(x^{n})$
    \end{algorithmic}
\end{algorithm}

Through the recursive operations in $\fpa$, features and the gradients could explode with increasing iterations and hence it can cause instabilities during training.
To avoid such a serious issue, we will follow the prescription employed in the original paper \cite{behrmann2019invertible} referred to as the spectral normalization,
which serves to stabilize the training process (and also inference) by normalizing the spectrum of the weight matrices.

Our update rule for the visible neurons, as well as the fixed-point iteration method performed in $\fpa$, can be regarded as a form of the deep equilibrium model or the implicit layer \cite{bai2019deep,ghaoui2021implicit}.
In such algorithms also, the solution converges to a fixed point, but they do not necessarily have a direct gradient descent method for training.

%% file: sec5.tex
\section{Experiments}
\label{sec:experiment}

In order to verify the validity of the \iblock ~module based on the correspondence with the Hopfield network side,
in this section we embed the module into MLP-Mixer, which we refer to as \imodel, and perform several experiments to compare our \imodel ~with the vanilla MLP-Mixer (Mixer).
In the main part of the experiments, we utilize the CIFAR-10 dataset \cite{cifar} to evaluate the performance of \imodel-Small (S), -Base (B) and -Large (L) models corresponding to the vanilla Mixer models.
We use PyTorch Image Models \texttt{timm} \cite{rw2019timm} to implement the models in all the experiments.%
\footnote{The code is available at \url{https://github.com/Toshihiro-Ota/imixer}.}
The experimental details are provided in \aref{appendix:details}.

\noindent\textbf{Network architecture.}
We incorporate the \iblock ~module \eref{eq:imlp} with the token-mixing blocks, and the rest of the \imodel ~has exactly the same structure as the vanilla Mixer.
That brings in the following additional hyperparameters to \imodel: the dimension of the hidden neurons $N_h$, the number of fixed-point iteration $\niter$,
and the number of power-iteration $n_p$ and the coefficient $c \, (<1)$ for the spectral normalization.
The large number of power-iteration $n_p$ stabilizes the training of the models with the i-Res module.
Based on the successful practices reported in \cite{behrmann2019invertible}, we will fix $n_p=8$ and $c=0.9$ in the experiments.
The effect of the spectral normalization will be discussed later in \sref{sec:ablation}.
As we take the dimension of the middle neurons equal to the usual spatial MLP dimension: $N_\xint = D_S$, the dimension of the hidden neurons as a hyperparameter is specified by the ratio to $D_S$,
that is, we have a hyperparameter $\hratio$ defined by $\hratio := N_h/D_S$.
In all, we have two hyperparameters in \imodel, $\hratio$ and $\niter$.
The pseudo-code of the \iblock ~module is provided in \aref{appendix:code}.

\subsection{CIFAR-10}
\label{sec:cifar10}

To evaluate the performance of our \imodel ~model, in this subsection we conduct thorough experiments on the scratch training of the model with CIFAR-10.
The statistics of the results are obtained by ten trials with random initialization through all the experiments, unless otherwise stated.
We also perform the hyperparameter search with \imodel-S, and with those results in hand, we will consider additional studies in the following subsections.

\noindent\textbf{Training.}
We use the CIFAR-10 dataset \cite{cifar}, which consists of 60,000 natural images of size $32\times 32$.
The ground-truth object category labels are attached to each image and the number of categories is 10, with 6,000 images per class.
There are 50,000 training images and 10,000 test images.
For the training setting, we basically follow the previous study \cite{TouvronICML2021}.
The images are resized to $224\times 224$ and the AdamW \cite{loshchilovdecoupled} optimizer is used.
We take the base learning rate $\frac{\text{batch size}}{512}\times 5\times 10^{-4}$.
The batch sizes for Small, Base and Large models are 512, 256 and 64, respectively.
As the regularization method, we employ label smoothing \cite{Szegedy_2016_CVPR} and stochastic depth \cite{huang2016deep}.
As for the data augmentation, we apply cutout \cite{devries2017improved}, cutmix \cite{Yun_2019_ICCV}, mixup \cite{zhangmixup}, random erasing \cite{zhong2020random}, and randaugment \cite{cubuk2020randaugment}.
For more details, see \aref{appendix:training}. We train \imodel ~and the vanilla Mixer in the exact same training configuration for fair comparison.
All the experiments are conducted with four Quadro RTX 8000 GPU cards.

\begin{table}[t]
  \centering
  \caption{Top-1 accuracy (\%) of \imodel ~($\hratio=2, \niter=2$), compared with the vanilla Mixer.}
  \label{tab:cifar10}
  \begin{tabular}{lccc}
    \toprule
            Model               &      Small                              &      Base                               &      Large  \\
    \midrule
    \phantom{i}Mixer (baseline) &      $88.08$  {\scriptsize $\pm 0.51$}  &      $89.03$  {\scriptsize $\pm 0.24$}  &      $86.67$  {\scriptsize $\pm 0.30$}  \\
            \imodel ~(ours)     &  $\bf{88.56}$ {\scriptsize $\pm 0.30$}  &  $\bf{89.07}$ {\scriptsize $\pm 0.33$}  &  $\bf{87.48}$ {\scriptsize $\pm 0.40$}  \\
    \bottomrule
  \end{tabular}
\end{table}

\noindent\textbf{Results.}
\tref{tab:cifar10} shows the results.
Our \imodel ~models consistently improve the performance of the baseline vanilla Mixer models.
The hyperparameters are taken as $\hratio=2$ and $\niter=2$.
The results for hyperparameter search is also provided in \tref{tab:hyper_hr_n}.
All the results in these tables are obtained from the scratch training on the models.
As for the prediction accuracy, one finds that the best hyperparameters are $\hratio=2$ and $\niter=2$, while $\niter=1$ already provides the competitive result.
We will make comments regarding this observation with connection to the residual networks below and in the next subsection.

\begin{table}[ht]
  \centering
  \caption{Hyperparameter search for $\hratio$ and $\niter$ in \imodel-S. Each slot represents the corresponding top-1 accuracy (\%).}
  \label{tab:hyper_hr_n}
  \begin{tabular}{cccc}
    \toprule
      $\hratio$ &      $\niter=1$                         &      $\niter=2$                         &      $\niter=4$  \\
    \midrule
      $0.25$    &      $88.26$  {\scriptsize $\pm 0.28$}  &      $88.22$  {\scriptsize $\pm 0.33$}  &      $88.29$  {\scriptsize $\pm 0.37$}  \\
      $0.5$     &      $88.32$  {\scriptsize $\pm 0.39$}  &      $88.21$  {\scriptsize $\pm 0.45$}  &      $88.22$  {\scriptsize $\pm 0.43$}  \\
      $1$       &      $88.36$  {\scriptsize $\pm 0.31$}  &      $88.32$  {\scriptsize $\pm 0.32$}  &      $88.32$  {\scriptsize $\pm 0.32$}  \\
      $2$       &  $\bf{88.54}$ {\scriptsize $\pm 0.34$}  &  $\bf{88.56}$ {\scriptsize $\pm 0.30$}  &  $\bf{88.46}$ {\scriptsize $\pm 0.26$}  \\
    \bottomrule
  \end{tabular}
\end{table}

From \tref{tab:cifar10}, we observe that the \imodel-L shows the largest improvement of the performance.
In addition, \tref{tab:hyper_hr_n} tells us that the larger the models become, the performance tends to be more improved from the vanilla Mixer models since the bigger $\hratio$ yields larger width in the hidden layers.
These observations are in a sense counterintuitive since, as MLP-Mixer (including our \imodel) has much less inductive bias than state-of-the-art neural networks such as CNN, the models with more parameters seem to get more overfitted with small training data, such as CIFAR-10.
Another aspect that can be observed in \tref{tab:cifar10} is the milder degradation of performance from Small to Large in our \imodel ~compared to the vanilla Mixer.
These counterintuitive behaviors imply that our \iblock ~module, derived from a Hopfield-type neural network, is effective.

\tref{tab:hyper_hr_n} does not show a significant statistical difference between $n=1$, $2$, and $4$.
This observation rather suggests that increasing the order $n$ within these experiments does not significantly affect the performance of the models.
In other words, while residual networks are indeed efficient neural network architectures, \imodel ~with larger $n=2$ and $4$ also exhibits competitive performance compared to $n=1$.
This implies the effectiveness of the i-Res module implementation in the proposed iMLP module and hence the validity of the formulation through the Hopfield/Mixer correspondence.

\begin{figure}[t]
    \centering
    \includegraphics[keepaspectratio, scale=0.4]{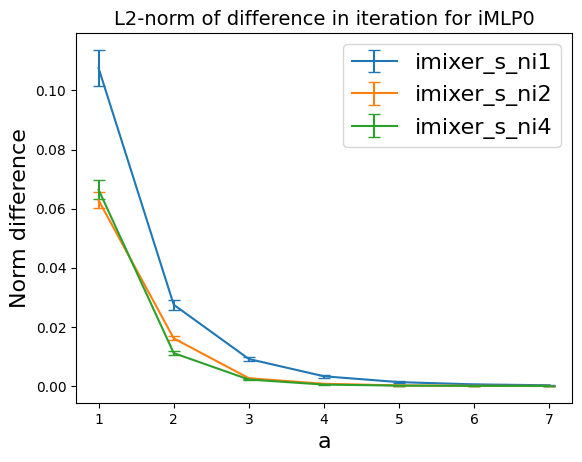}
    \hspace{1em}
    \includegraphics[keepaspectratio, scale=0.4]{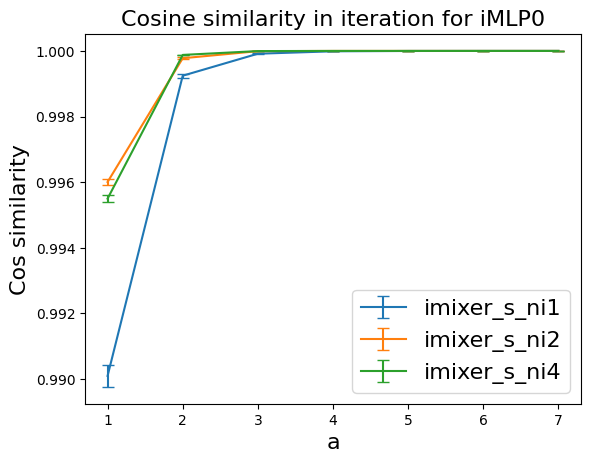}
    \caption{Convergence rate of $L_2$-norm (left), $\| x^{a+1}-x^a \|$, and cosine similarity (right), $\cos(x^{a+1},x^a)$,
            between two successive feature vectors in fixed-point iteration $\fpa$ in the \iblock ~module of the first layer (iMLP-0).}
    \label{fig:imlp_iter}
\end{figure}

Nevertheless, $n=1$ generically does not suffice for the accuracy of the expansion approximation $\fpa$, \eref{eq:fpamodule}, in the \iblock ~module.
Figure \ref{fig:imlp_iter} shows $L_2$-norm and cosine similarity between two successive feature vectors $x^{a+1}$ and $x^a$ in iteration.
The inputs are $16$ test samples of CIFAR-10 chosen at random, and the $L_2$-norm is normalized by the internal dimensions due to its scaling.
The model names \texttt{imixer\_s\_ni*} indicate the trained \imodel-S models with $\hratio=2$ and the number of fixed-point iteration $n=1$, $2$ and $4$, respectively.
We observe that the models \texttt{imixer\_s\_ni2} and \texttt{imixer\_s\_ni4} yield approximately same convergence rate and $n=2$ or $4$ suffice for the expansion approximation, whereas $n=1$ is not enough.
We here show the result of the first layer in \imodel, \iblock-0.
The plots corresponding to \iblock ~modules of other layers are given in \aref{appendix:imlp_iteration}.

\subsection{Ablation study}
\label{sec:ablation}

\begin{table}[t]
  \centering
  \caption{Ablation study for the spectral normalization in \imodel-S ($\hratio=2$), trained on CIFAR-10. Each slot represents the corresponding top-1 accuracy (\%).}
  \label{tab:specnorm}
  \begin{tabular}{lccc}
    \toprule
      Norm         & $\niter=1$                         & $\niter=2$                         & $\niter=4$  \\
    \midrule
      SpecNorm     & $88.54$  {\scriptsize $\pm 0.34$}  & $88.56$  {\scriptsize $\pm 0.30$}  & $88.46$  {\scriptsize $\pm 0.26$}  \\
      w/o SpecNorm & $88.73$  {\scriptsize $\pm 0.39$}  & $88.32$  {\scriptsize $\pm 0.31$}  & $88.07$  {\scriptsize $\pm 0.28$}  \\
      BatchNorm    & $87.60$  {\scriptsize $\pm 0.34$}  & $81.62$  {\scriptsize $\pm 1.98$}  & $81.83$  {\scriptsize $\pm 2.00$}  \\
    \bottomrule
  \end{tabular}
\end{table}

As already mentioned, the \iblock ~module actually involves the spectral normalization to make the backward operation $(1-F)^{-1}$ a usual forward network as in \eref{eq:imlp}.
We study the effect of the spectral normalization by replacing the normalization condition, instead of changing the parameters $n_p$ and $c$.
While searching for optimal $n_p$ and $c$ could be beneficial, it appears to deviate from our paper's primary objective.
From \tref{tab:specnorm}, one can see that the replacement of the spectral normalization (SpecNorm) with the batch normalization (BatchNorm) apparently made worse the prediction accuracy of the model,
where for the $\niter=4$, Norm=BatchNorm case, we reduced the batch size because we encountered a memory issue when the original batch size is used in our environment.
One may find from \tref{tab:specnorm} that for $\niter=1$ the model performance seems to slightly improve when the spectral normalization is removed (w/o SpecNorm).
This is actually not the case for $\niter=2$ and $4$, rather larger $\niter$ makes the performance worse.
This tendency is consistent with the role of the spectral normalization in the i-Res module.
This indicates that the spectral normalization is inevitable for the fixed-point iteration taken in the \iblock ~module, while the $\fpa$ turns out to be a naive residual connection for the $\niter=1$ case.
Thus, we can conclude that the spectral normalization is necessary for the assurance of the performance.

\subsection{Other datasets}

\begin{table}[t]
  \centering
  \caption{First three lines show sample size and the number of classes for each dataset.
          The last two rows show the top-1 accuracy (\%) of the vanilla Mixer-S and \imodel-S ~($\hratio=2, \niter=2$) trained on the corresponding dataset.
          Ten runs for each, except for ImageNet-1k.}
  \label{tab:datasets_result}
  \begin{tabular}{lcccc}
    \toprule
                        & CIFAR-100                          & Stanford Cars                     & Food-101                           & ImageNet-1k  \\
    \midrule
    Train size          & 50,000                             & 8,144                             & 75,750                             & 1,281,167    \\
    Test size           & 10,000                             & 8,041                             & 25,250                             & 50,000       \\
    \#Classes           & 100                                & 196                               & 101                                & 1,000        \\
    \cmidrule(rl){1-5}
    \phantom{i}Mixer-S  & $68.13$  {\scriptsize $\pm 0.46$}  & $8.09$  {\scriptsize $\pm 0.45$}  & $76.11$  {\scriptsize $\pm 0.32$}  & $73.91$      \\
              \imodel-S & $68.26$  {\scriptsize $\pm 0.30$}  & $7.96$  {\scriptsize $\pm 0.17$}  & $76.08$  {\scriptsize $\pm 0.20$}  & $74.10$      \\
    \bottomrule
  \end{tabular}
\end{table}

We also investigate the effectiveness of our \imodel ~for other datasets.
We here utilize the commonly used datasets such as CIFAR-100 \cite{cifar}, Stanford Cars \cite{KrauseStarkDengFei-Fei_3DRR2013}, Food-101 \cite{bossard14}, and ImageNet-1k \cite{NIPS2012_c399862d}.
\tref{tab:datasets_result} shows the results of scratch training on the vanilla Mixer-S and \imodel-S models, and also provides the details on the datasets.
The training configuration is taken exactly the same as in the previous subsections.
From this table, one finds that our \imodel ~models show reasonably the competitive performance compared to the corresponding vanilla Mixer models.
An exception is for the Stanford Cars dataset, which is composed of only 8,144 training samples and 8,041 test samples labeled with 196 classes.
This result indicates that the scratch training both on the vanilla Mixer and on \imodel ~failed for such a too small sample size with relatively large number of classes.
\tref{tab:datasets_result} shows that \imodel ~exhibits the competitive performance as a novel model of MetaFormer, which is consistent with the results shown in \tref{tab:cifar10}.

%% file: sec6.tex
\section{Discussion}

\subsection{Limitation}

One limitation is the lack of application of iMixer for other computer vision tasks, such as segmentation, anomaly detection, and robustness, while including those discussions may make our main contributions obscure.

Another limitation is seen in the fixed number of hidden layers in our formulation of \imodel, in view of the \kro.
In this paper we have only considered the case of three layers in total, while the formulation of \imodel ~can be generalized to any number of layers as given in \aref{appendix:general}.
To include more hidden layers would give us more supports and variations for the correspondence between the Hopfield networks and the \imodel.

A narrow focus on the Lagrangians in the formulation is another limitation.
The Lagrangians for hidden neurons (in other words, activation functions $f$ and $e$ in \sref{sec:Derivation}) are actually not restricted along the derivation of \imodel, where we fixed $f=e=\gelu$ just for practical reasons.
Reconsideration of the Lagrangians for hidden neurons might give some more insights in understanding of the role of token mixers in MetaFormers.
We leave these aspects of study for future works.

\subsection{Conclusion}

We proposed a generalized Hopfield/Mixer correspondence and found \imodel, invertible, implicit and iterative MLP-Mixer, as the counterpart of the hierarchical modern Hopfield network.
The proposed \iblock ~module is an example of implicit layer that includes the deep equilibrium model.
Although the primary interest of this paper is not intended to improve the accuracy of the state-of-the-art architecture, but to give a unified description for architecture design of MetaFormers such as MLP-Mixer, we also provided the empirical evaluation.
The evaluation experiments showed that \imodel ~performs reasonably well compared to the baselines and that the proposed model works.

\imodel ~involves MLP layers that propagate forward from the output side to the input side.
Although this architectural design of \imodel ~may seem unconventional from a computer vision perspective, the empirical studies show that the larger the models become, the performance tends to be more improved from the vanilla Mixer models.
Despite having less inductive bias than state-of-the-art neural networks like CNN, \imodel ~exhibits a somewhat counterintuitive trend where larger models with more parameters tend to perform better, even with limited training data that could potentially cause overfitting.
This counterintuitive behavior suggests that exploring the correspondence between Hopfield networks and Mixer models could be a promising direction for describing MetaFormers more comprehensively.

%% file: appendix.tex
\section{Broader Impact}

Our work provides a variant of MetaFormers through the update rule of the hierarchical modern Hopfield network equipped with the corresponding Lagrangians.
From this point of view, we can further study the fundamental properties of the token-mixing modules in state-of-the-art models by mapping them to the dynamics of Hopfield networks, which are more transparent neural network models.
In addition, this perspective enables us to explore more efficient and generalizable MetaFormer models in a transparent way, not in ad hoc ways from tasks to tasks.

Since our experiments were designed to verify the formulation of our proposed \imodel,
we believe that our results do not directly cause harm to society, while they can be of use to develop new Transformer-type architectures.
It is expected that a more promising Hopfield model with a set of Lagrangians and the associated energy functions will be a good starting point to pursue brand-new neural network models.

\section{A General Formulation of \imodel}
\label{appendix:general}

In this appendix, we describe one of the most general formulations of the \imodel.
We begin with the setup of Krotov's hierarchical Hopfield network \cite{krotov2021hierarchical} with $L$ layers in total, which consists of $N_A$ ($A=1,\dots,L$) neurons in each layer.
This system is a straightforward generalization of the model discussed in \sref{sec:model} and basically given by the following components:
$x^A(t)\in \R^{N_A}$ neurons at each layer, interaction matrices between the adjacent layers, $\xii{A}{A-1}\in \R^{N_A\times N_{A-1}}$, and activation functions
\begin{equation}
  g^A: \R^{N_A} \to \R^{N_A},  \quad  A = 1, \dots, L,
\end{equation}
determined through Lagrangians $L^A: \R^{N_A}\to \R$ such that $g^A(x^A) = \partial L^A(x^A)/\partial x^A$.
For more details, see \cite[Sec.~3]{krotov2021hierarchical}.
With these notations, the dynamical equations describing the system of neurons are
\begin{equation}
  \tau_A\dd{x^A(t)}{t}
  = \xii{A}{A-1}g^{A-1}\left(x^{A-1}(t)\right) + \xii{A}{A+1}g^{A+1}\left(x^{A+1}(t)\right) - x^A(t),
\end{equation}
with boundary conditions $g^0\equiv 0$ and $g^{L+1}\equiv 0$.
One can easily find that by taking $L=3$ and seeing that $x^1=v$, $x^2=x$ and $x^3=h$, the system produces the one discussed in the main text.
In this appendix, we always have in mind that $x^1$ neurons only are the visible ones and $x^L$ are the lowest hidden ones.

\begin{figure}[t]
  \centering
  \includegraphics[keepaspectratio, scale=0.5]{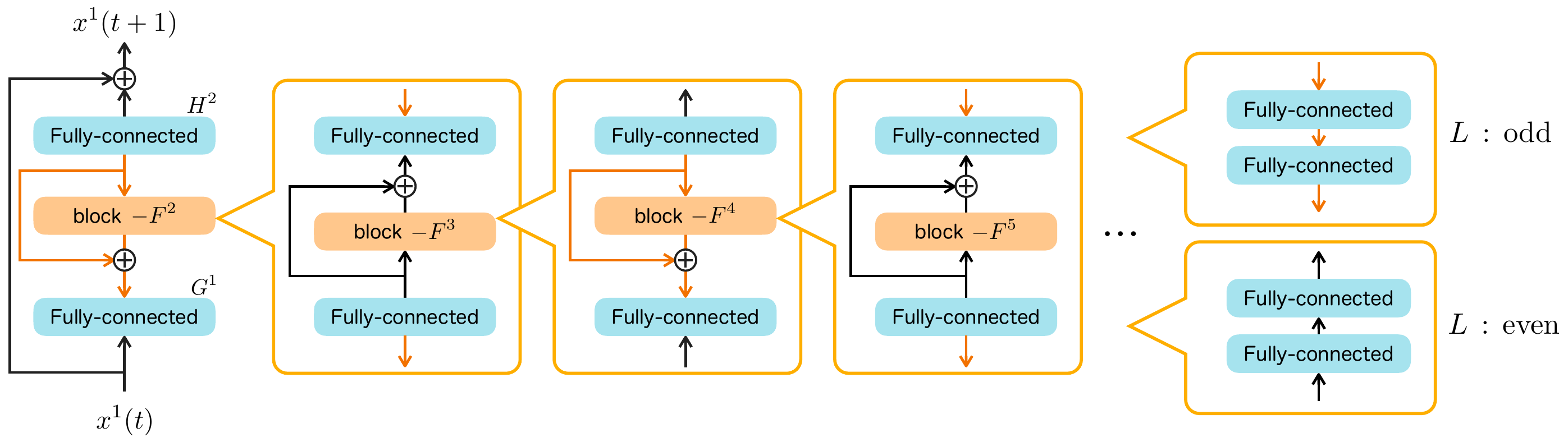}
  \caption{The structure of \iblock ~module for generic $L$.}
  \label{fig:fig_hierarchical_imixer}
\end{figure}

We here again consider the addiabatic limit, $\tau_1 \gg \tau_2 \gg \cdots \gg \tau_L$,
which means that the dynamics of the hidden neurons in lower layers is always much faster than that of in the higher layers.
Then, the dynamical equations turn out to be
\begin{align}
  \tau_1\dd{x^1(t)}{t}
    &= \xii{1}{2}g^{2}\left(x^{2}(t)\right) - x^1(t),  \\
  x^A(t)
    &= \xii{A}{A-1}g^{A-1}\left(x^{A-1}(t)\right) + \xii{A}{A+1}g^{A+1}\left(x^{A+1}(t)\right),  \\
  x^L(t)
    &= \xii{L}{L-1}g^{L-1}\left(x^{L-1}(t)\right),
\end{align}
where $A$ in the middle equation runs from $2$ to $L-1$.
Now we take the Lagrangian for the $x^1$ neurons as
\begin{equation}
  L^1(x^1) = \sqrt{ \sum_{i=1}^{N_1} (x^1_i - \bar{x}^1)^2 },
  \quad
  \bar{x}^1 = \frac{1}{N_1} \sum_i x^1_i.
\end{equation}
Other Lagrangians for the hidden neurons actually need not be specified to derive the general formulation of \imodel.
By following the same discussion as in \sref{sec:Mixer} in the main text, the dynamical equation for the $x^1$ neurons eventually becomes
\begin{equation}
    x^1(t+1) = x^1(t) + H^2(x^2(t)),
    \quad
    H^2 := \xii{1}{2} g^2(\cdot).
\end{equation}
To solve this equation, we need a solution for the $x^2$ neurons, which can be obtained by the i-Res module method as follows.
Using the above reduced dynamical equations from $A=L$ to $A=2$ one-by-one, one eventually obtains the solution as
\begin{equation}
  x^2(t) = \left( 1 - F^2 \right)^{-1} \left( G^1 (x^1(t)) \right),  \label{eq:sol-x2}
\end{equation}
where $F^A: \R^{N_A} \to \R^{N_A}$ are contractive MLP blocks defined as
\begin{align}
  F^A &= H^{A+1} \circ \left( 1 - F^{A+1} \right)^{-1} \circ G^A,
  \quad
  A = 1, \dots, L-2,  \\
  F^{L-1} &= H^{L} \circ G^{L-1}.
\end{align}
Fully-connected layers $G^A: \R^{N_A} \to \R^{N_{A+1}}$ and $H^{A+1}: \R^{N_{A+1}} \to \R^{N_A}$ are defined by
\begin{equation}
  G^A = \xii{A+1}{A} g^{A},  \quad  H^{A+1} = \xii{A}{A+1}g^{A+1}.
\end{equation}
Thus, by substituting \eref{eq:sol-x2} to the discrete dynamical equation for the $x^1$ neurons, we obtain
\begin{equation}
  x^1(t+1) = x^1(t) + F^1(x^1(t)),
\end{equation}
with a hidden series of contractive MLP blocks $F^A$ given as above; see \fref{fig:fig_hierarchical_imixer}.
By utilizing the i-Res module method to the each $(1-F^A)^{-1}$ and replacing the inverse operations with the fixed-point approximations $\fpa^A$, we have a general \imodel ~architecture.
Taking $L=3$, one finds that this update rule for the $x^1$ neurons reproduces our \iblock ~module \eref{eq:imlp}.

\newpage
\section{Experimental Details}
\label{appendix:details}

This section provides the experimental details, including the feature maps of the \iblock ~module.

\subsection{Code of the \iblock ~module}
\label{appendix:code}

The pseudo-code of the \iblock ~module is shown in Algorithm \ref{alg:pseudocode}.
As mentioned in the main text, we use PyTorch Image Models \texttt{timm} \cite{rw2019timm}%
\footnote{%
  \url{https://github.com/huggingface/pytorch-image-models}.
}
for the implementation of the models.
The \imodel ~models mostly have the same structure as the vanilla MLP-Mixer models, except for the use of this \iblock ~module for the token-mixing block.
The module \texttt{spectral\_norm\_fc} in the code is the spectral normalization discussed and provided in \cite{behrmann2019invertible}.%
\footnote{%
  \url{https://github.com/jhjacobsen/invertible-resnet}.
}

\begin{algorithm}
\caption{Pseudo-code of the invertible MLP layer, PyTorch-like code.}
\label{alg:pseudocode}
\begin{lstlisting}[language=Python]
# MLP block with the invertible ResNet module
class iMlp(nn.Module):
    def __init__(self, d_vis, d_mid, d_hid, act=nn.GELU, drop=0.,
                coeff=0.9, n_power=8, n_iter=2):
        super().__init__()
        self.fc1 = nn.Linear(d_vis, d_mid)
        self.fc_sn1 = spectral_norm_fc(nn.Linear(d_mid, d_hid),
                                        coeff, n_power)
        self.fc_sn2 = spectral_norm_fc(nn.Linear(d_hid, d_mid),
                                        coeff, n_power)
        self.fc2 = nn.Linear(d_mid, d_vis)

        self.act = act()
        self.drop = nn.Dropout(drop)
        self.n_iter = n_iter

    # feedforward operation with the i-Res module:
    # the fixed-point iteration method
    def forward(self, x):
        x = self.fc1(x)
        x_in = x
        for _ in range(self.n_iter):
            x = self.act(x)
            x = self.drop(x)
            x = self.fc_sn1(x)
            x = self.act(x)
            x = self.drop(x)
            x = self.fc_sn2(x)
            x = x + x_in
        x = self.act(x)
        x = self.drop(x)
        x = self.fc2(x)
        x = self.drop(x)
        return x
\end{lstlisting}
\end{algorithm}

\newpage
\subsection{Training details}
\label{appendix:training}

We here report the detailed training setups commonly used for the vanilla MLP-Mixer and \imodel ~in \tref{tab:hyper0}.
We basically follow the previous study \cite{TouvronICML2021}, and also employ \cite{hou2022vision}%
\footnote{%
  \url{https://github.com/houqb/VisionPermutator}.
}
for some considerations.
The slashes indicate that the corresponding parameters are taken to the Small, Base, and Large models, respectively.

\begin{table}[ht]
\centering
\caption{Hyperparameters commonly used for the vanilla MLP-Mixer and our \imodel ~for fair comparison.}
\label{tab:hyper0}
\begin{tabular}{lc}
    \toprule
    Training configuration &  Small/Base/Large  \\
    \midrule
    optimizer              &  AdamW  \\
    training epochs        &  300  \\
    batch size             &  512/256/64  \\
    base learning rate     &  5e-4/2.5e-4/6.25e-5  \\
    weight decay           &  0.05  \\
    optimizer $\epsilon$   &  1e-8  \\
    optimizer momentum     &  $\beta_1=0.9$, $\beta_2=0.999$  \\
    learning rate schedule &  cosine decay  \\
    lower learning rate bound & 1e-6  \\
    warmup epochs          &  20  \\
    warmup schedule        &  linear  \\
    warmup learning rate   &  1e-6  \\
    cooldown epochs        &  10  \\
    crop ratio             &  0.875  \\
    RandAugment            &  (9, 0.5)  \\
    mixup $\alpha$         &  0.8  \\
    cutmix $\alpha$        &  1.0  \\
    random erasing         &  0.25  \\
    label smoothing        &  0.1  \\
    stochastic depth       &  0.1/0.2/0.3  \\
    \bottomrule
\end{tabular}
\end{table}

\newpage
\subsection{Convergence rate for fixed-point iteration in \iblock ~modules}
\label{appendix:imlp_iteration}

We here show the convergence rate for fixed-point iterations in a trained \imodel-S, which has eight \iblock ~modules in total.
The setup is exactly the same as in \fref{fig:imlp_iter} in the main text; we consider $L_2$-norm and cosine similarity between two successive feature vectors $x^{a+1}$ and $x^a$ of $\fpa$ in each \iblock ~module:
$\mathrm{Norm}_a := \| x^{a+1} - x^a \|/\sqrt{S C}$ and $\mathrm{Cos}_a := \cos(x^{a+1}, x^a)$, where $S=196$ and $C=512$.
A point is that the convergence rate of the fixed-point iteration method depends on samples and weights, which means in general the results differ samples by samples and layer by layer.

Below we provide the results for convergence rate of fixed-point iteration in each \iblock ~module.
The inputs are 16 test samples of CIFAR-10.
From these plots, we see that the first two layers show the clear difference between \texttt{imixer\_s\_ni1} and \texttt{imixer\_s\_ni2} or \texttt{imixer\_s\_ni4}, due to the detail of the input data.
$\fpa$s in middle layers converge relatively fast, probably since the middle layers mainly manipulate abstract features of the data, independent of the details.
The final layer shows the large deviations, which implies the convergence rate differs for the classes sample by sample.
In all, this results suggest it is challenging and unreasonable to completely determine the convergence rate of the \iblock ~modules.

\begin{figure}[ht]
  \begin{tabular}{cccc}
  \begin{minipage}[b]{0.2475\linewidth}
    \centering
    \includegraphics[keepaspectratio, scale=0.22]{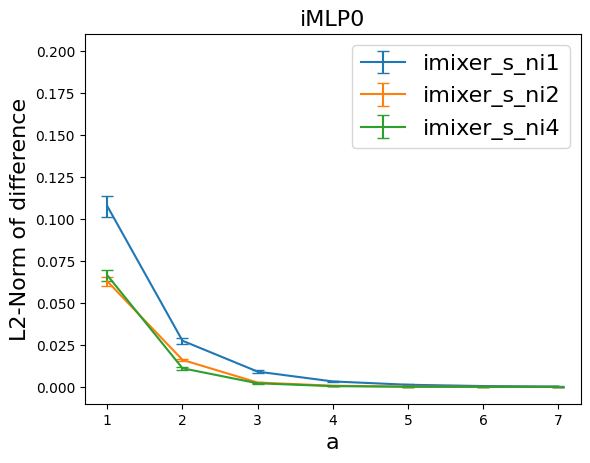}
  \end{minipage}
  \begin{minipage}[b]{0.2475\linewidth}
    \centering
    \includegraphics[keepaspectratio, scale=0.22]{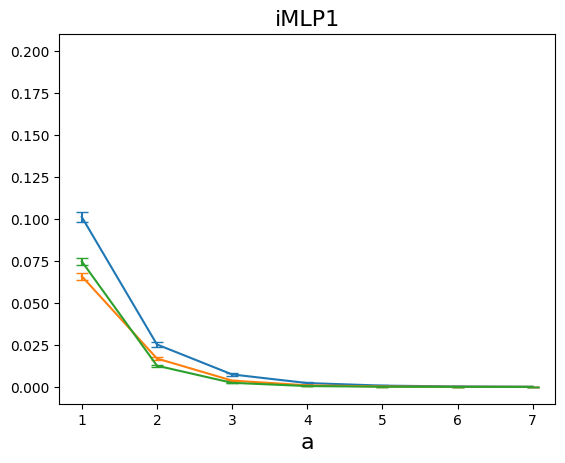}
  \end{minipage}
  \begin{minipage}[b]{0.2475\linewidth}
    \centering
    \includegraphics[keepaspectratio, scale=0.22]{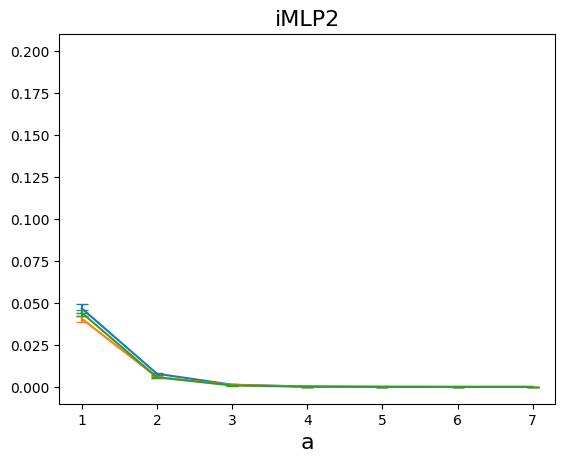}
  \end{minipage}
  \begin{minipage}[b]{0.2475\linewidth}
    \centering
    \includegraphics[keepaspectratio, scale=0.22]{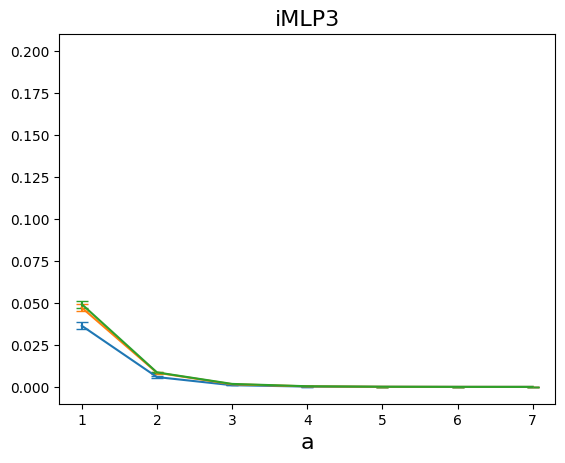}
  \end{minipage} \\
  \begin{minipage}[b]{0.2475\linewidth}
    \centering
    \includegraphics[keepaspectratio, scale=0.22]{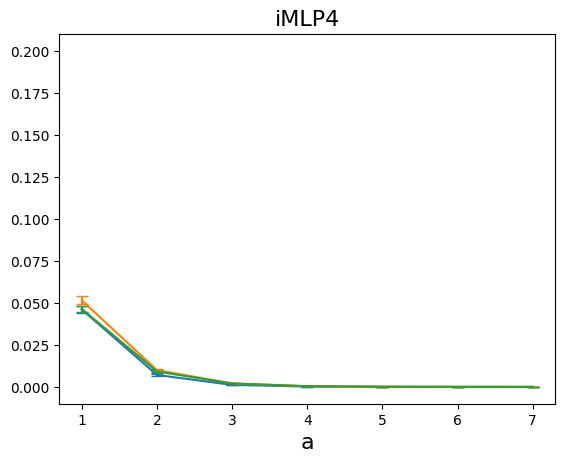}
  \end{minipage}
  \begin{minipage}[b]{0.2475\linewidth}
    \centering
    \includegraphics[keepaspectratio, scale=0.22]{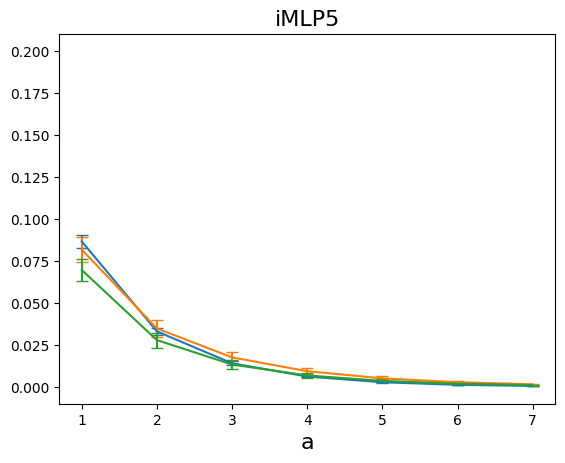}
  \end{minipage}
  \begin{minipage}[b]{0.2475\linewidth}
    \centering
    \includegraphics[keepaspectratio, scale=0.22]{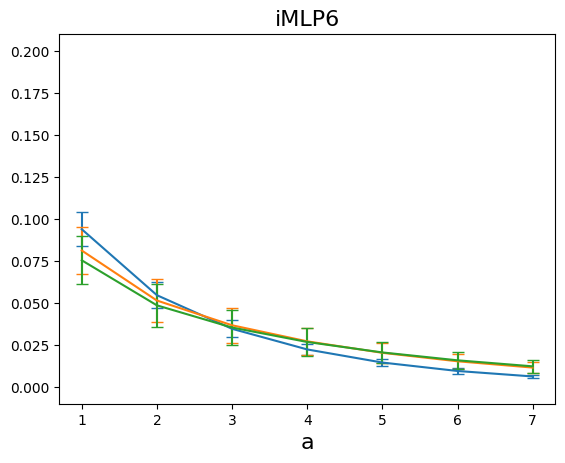}
  \end{minipage}
  \begin{minipage}[b]{0.2475\linewidth}
    \centering
    \includegraphics[keepaspectratio, scale=0.22]{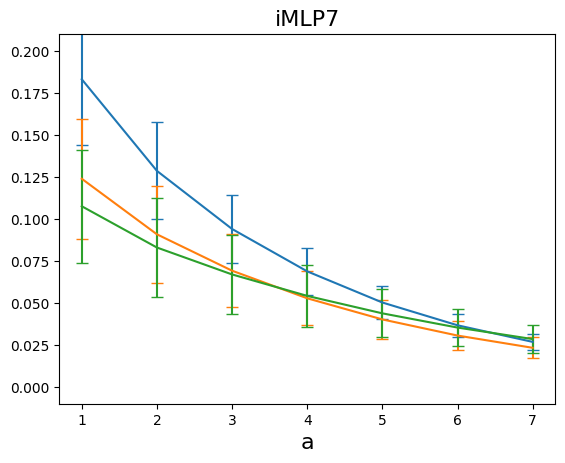}
  \end{minipage}
  \end{tabular}
  \caption{$L_2$-norm differences in fixed-point iteration for \iblock ~modules. The vertival axis is $\mathrm{Norm}_a$.}
\end{figure}

\begin{figure}[ht]
  \begin{tabular}{cccc}
  \begin{minipage}[b]{0.2475\linewidth}
    \centering
    \includegraphics[keepaspectratio, scale=0.22]{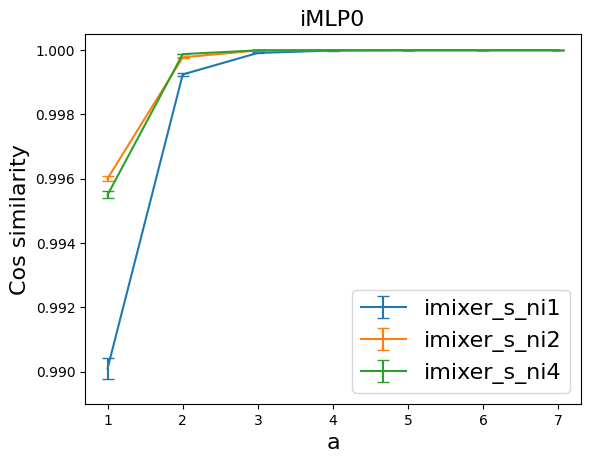}
  \end{minipage}
  \begin{minipage}[b]{0.2475\linewidth}
    \centering
    \includegraphics[keepaspectratio, scale=0.22]{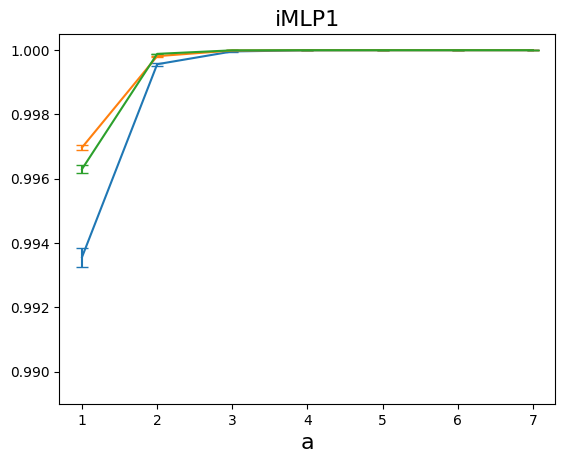}
  \end{minipage}
  \begin{minipage}[b]{0.2475\linewidth}
    \centering
    \includegraphics[keepaspectratio, scale=0.22]{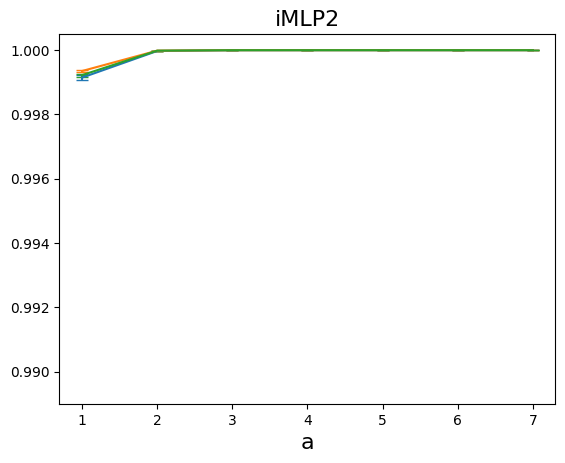}
  \end{minipage}
  \begin{minipage}[b]{0.2475\linewidth}
    \centering
    \includegraphics[keepaspectratio, scale=0.22]{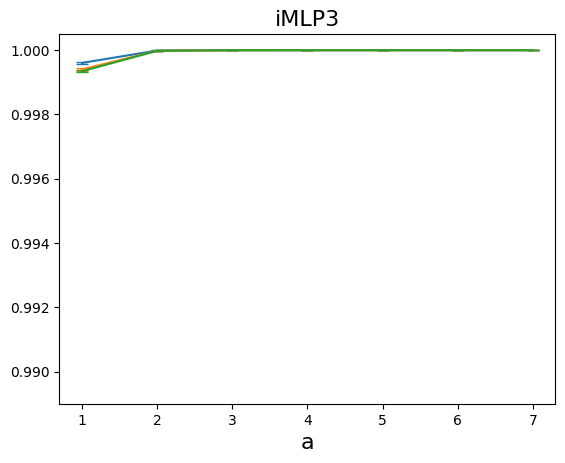}
  \end{minipage} \\
  \begin{minipage}[b]{0.2475\linewidth}
    \centering
    \includegraphics[keepaspectratio, scale=0.22]{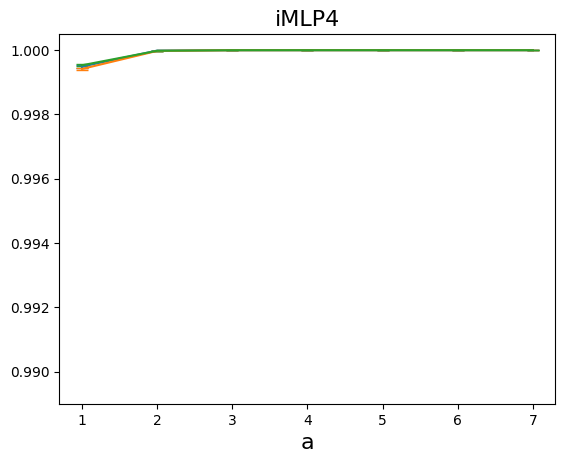}
  \end{minipage}
  \begin{minipage}[b]{0.2475\linewidth}
    \centering
    \includegraphics[keepaspectratio, scale=0.22]{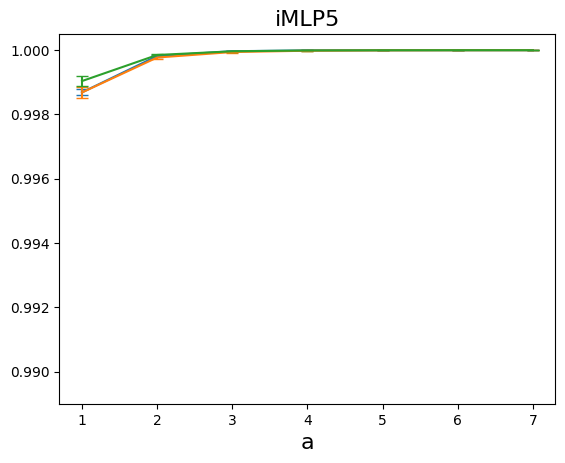}
  \end{minipage}
  \begin{minipage}[b]{0.2475\linewidth}
    \centering
    \includegraphics[keepaspectratio, scale=0.22]{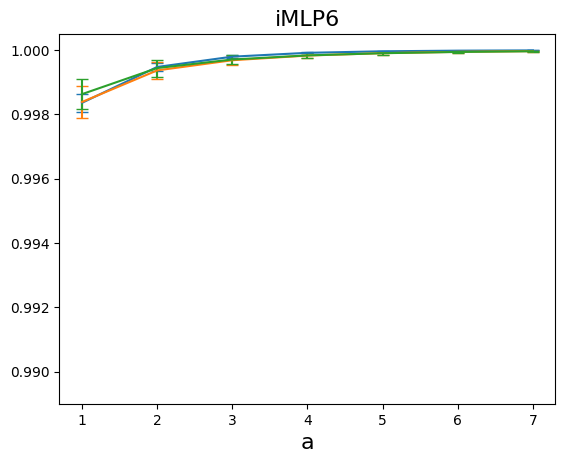}
  \end{minipage}
  \begin{minipage}[b]{0.2475\linewidth}
    \centering
    \includegraphics[keepaspectratio, scale=0.22]{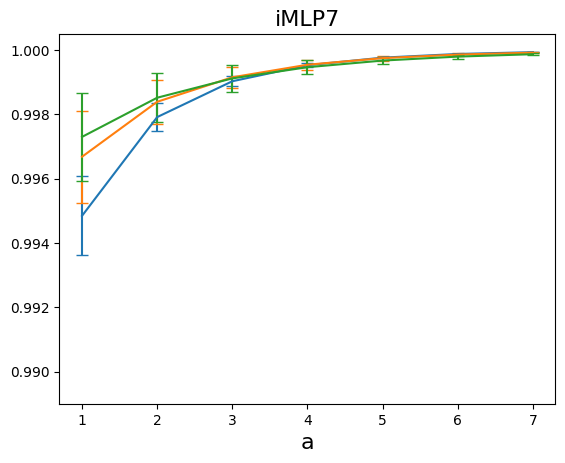}
  \end{minipage}
  \end{tabular}
  \caption{Cosine similarities in fixed-point iteration for \iblock ~modules. The vertival axis is $\mathrm{Cos}_a$.}
\end{figure}

\newpage
\subsection{Feature maps}
\label{appendix:featuremap}

We provide a specific visualization of the feature maps of the fixed-point iteration below.

\begin{figure}[ht]
  \begin{tabular}{cc}
    \begin{minipage}[b]{0.49\linewidth}
        \centering
        \includegraphics[keepaspectratio, scale=0.15]{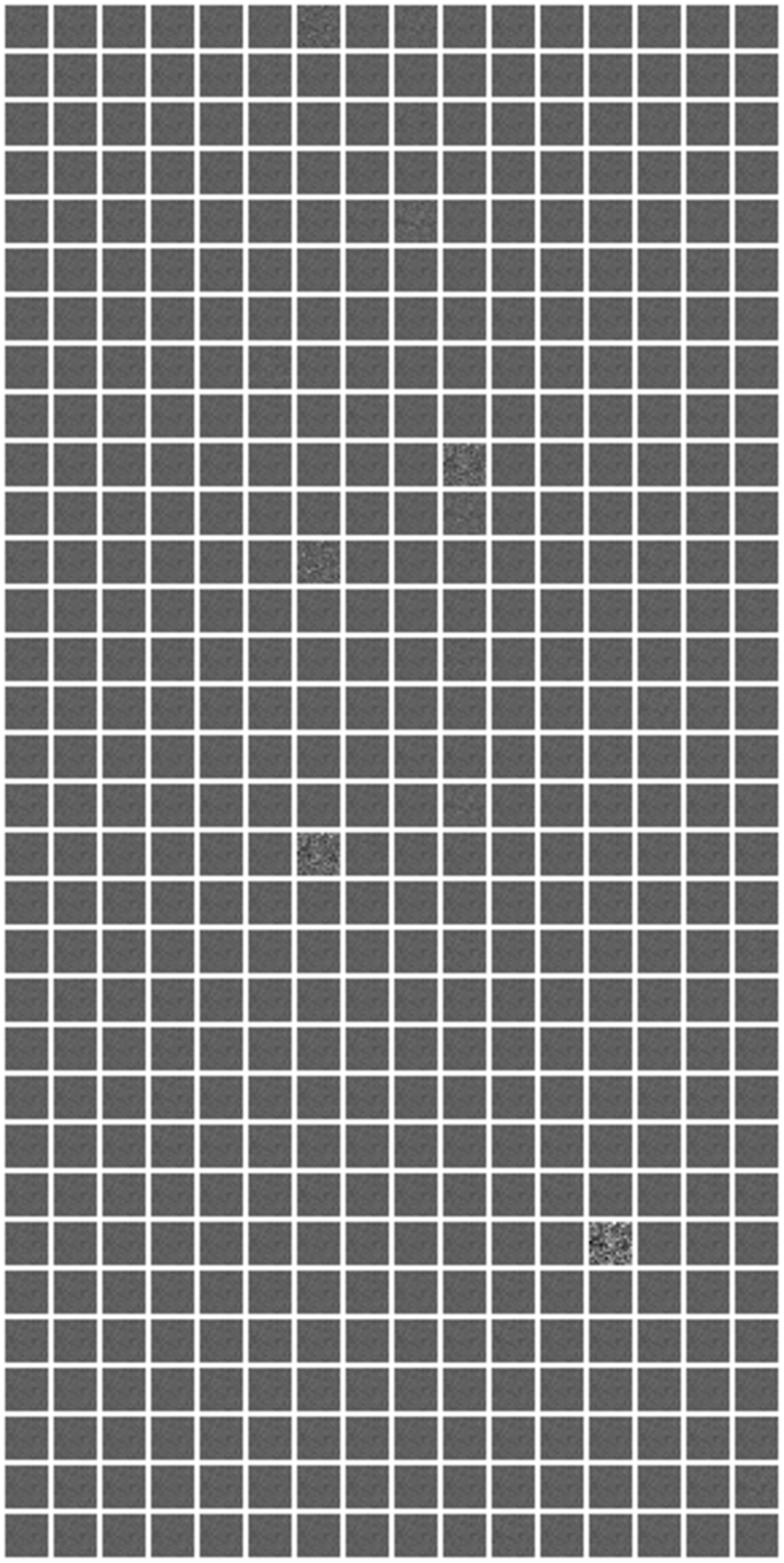}
        \subcaption{At first iteration.}
    \end{minipage}
    \begin{minipage}[b]{0.49\linewidth}
        \centering
        \includegraphics[keepaspectratio, scale=0.15]{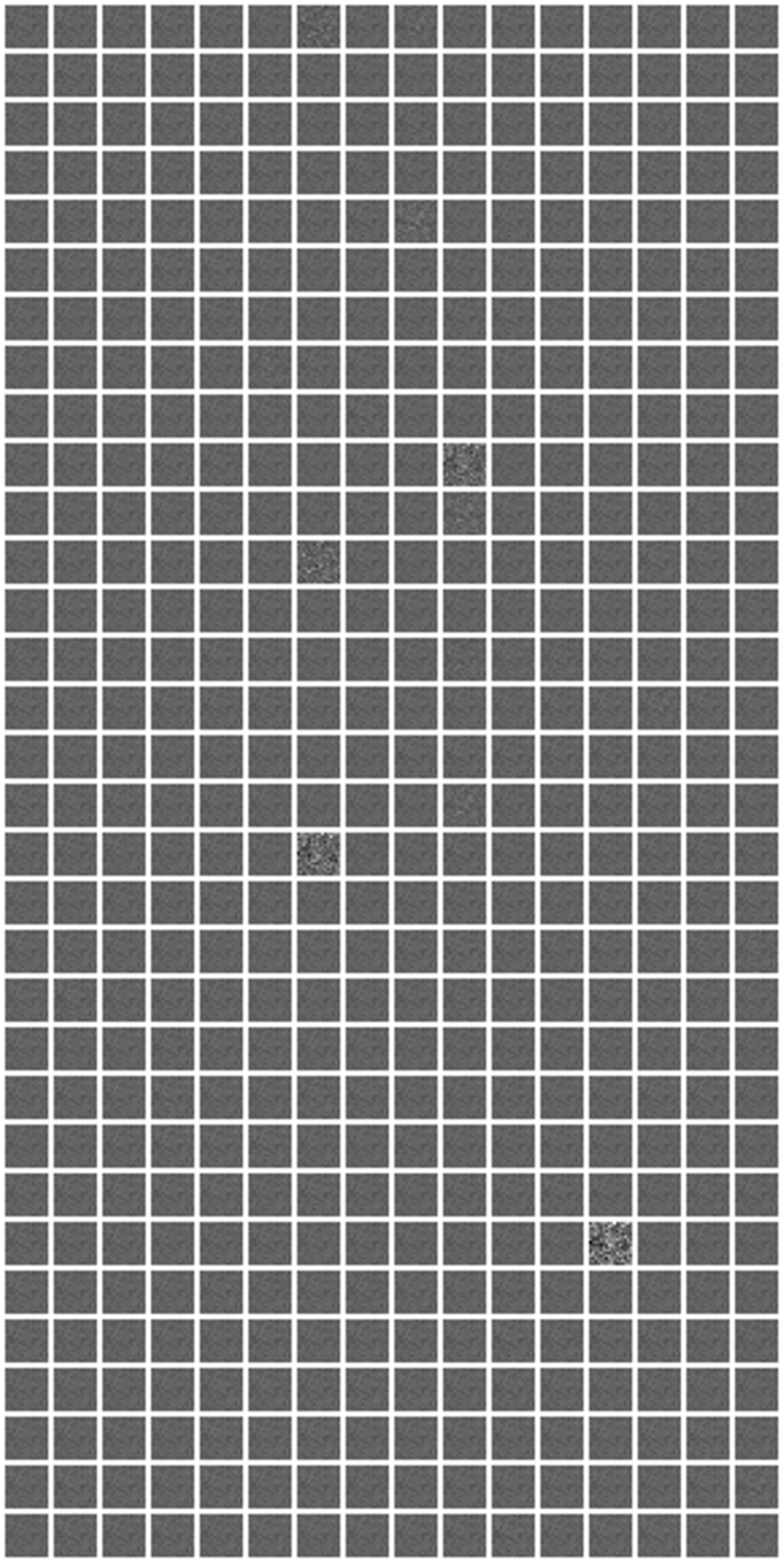}
        \subcaption{At second iteration.}
    \end{minipage} \\
    \begin{minipage}[b]{0.49\linewidth}
        \centering
        \includegraphics[keepaspectratio, scale=0.15]{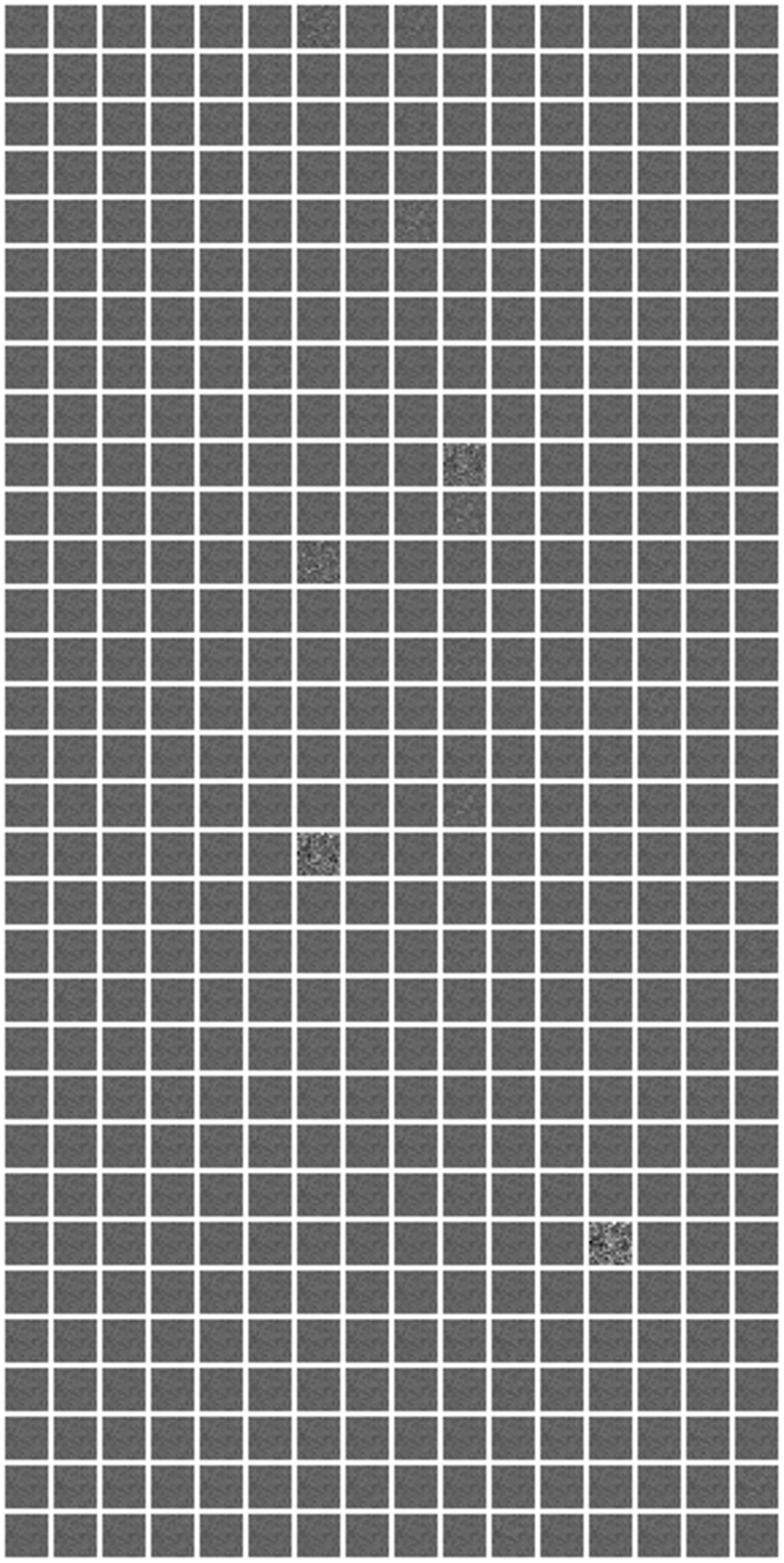}
        \subcaption{At third iteration.}
    \end{minipage}
    \begin{minipage}[b]{0.49\linewidth}
        \centering
        \includegraphics[keepaspectratio, scale=0.15]{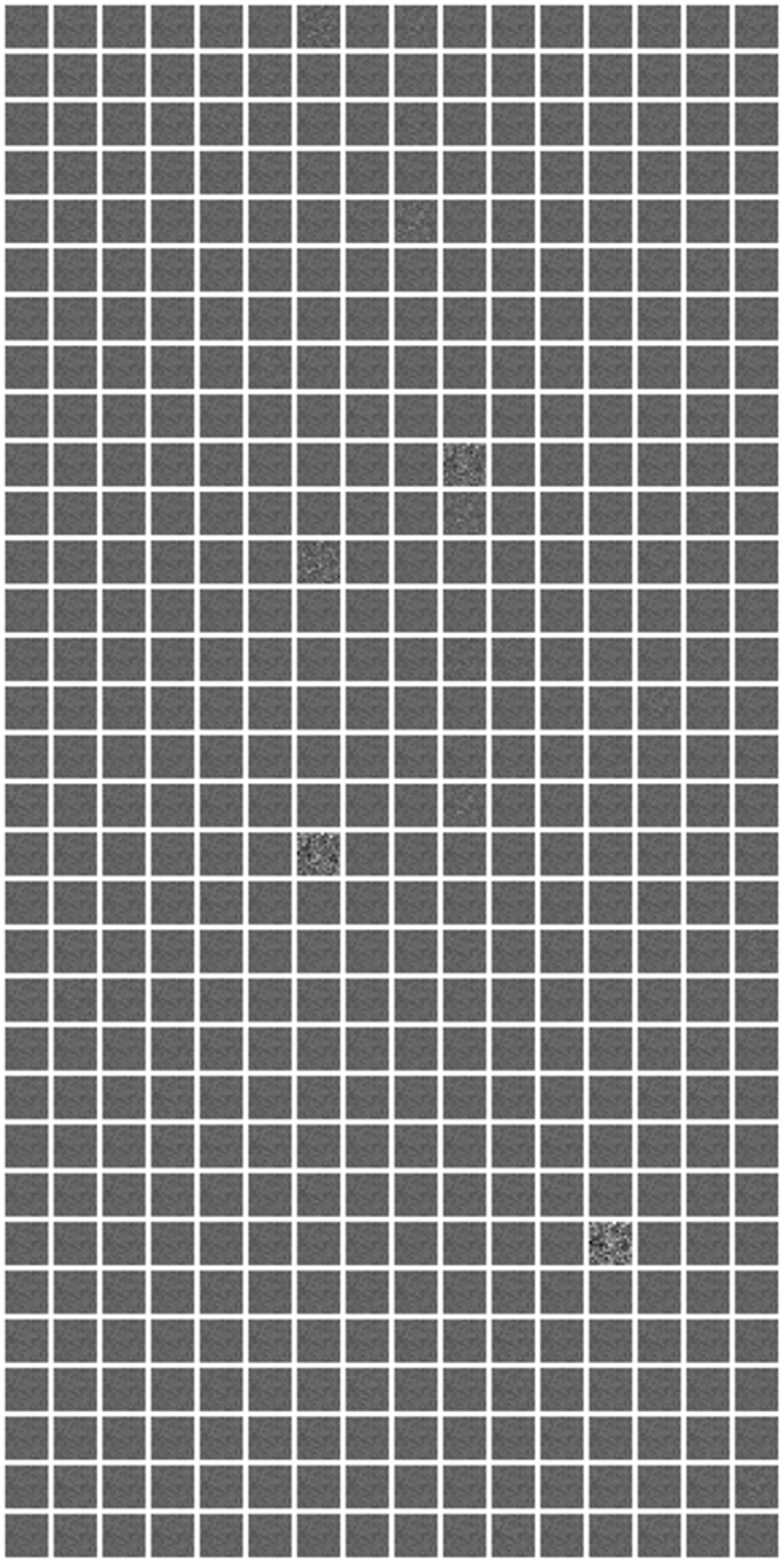}
        \subcaption{At fourth (final) iteration.}
    \end{minipage}
  \end{tabular}
  \caption{Feature maps (outputs) of each fixed-point iteration at the final layer of iMixer-S with $h_r=2$ and $n=4$ trained on ImageNet-1k.
          An input image is a test sample of dog.}
\end{figure}

%% file: main.bbl
\begin{thebibliography}{10}

\bibitem{NIPS2017_3f5ee243}
Vaswani, A., N.~Shazeer, N.~Parmar, et~al.
\newblock Attention is all you need.
\newblock In \emph{Advances in Neural Information Processing Systems}, vol.~30.
  2017.

\bibitem{DosovitskiyICLR2021}
Dosovitskiy, A., L.~Beyer, A.~Kolesnikov, et~al.
\newblock {An Image is Worth 16x16 Words: Transformers for Image Recognition at
  Scale}.
\newblock In \emph{International Conference on Learning Representation}. 2021.

\bibitem{TouvronICML2021}
Touvron, H., M.~Cord, M.~Douze, et~al.
\newblock {Training data-efficient image transformers \& distillation through
  attention}.
\newblock In \emph{International Conference on Machine Learning}. 2021.

\bibitem{tolstikhin2021mlp}
Tolstikhin, I.~O., N.~Houlsby, A.~Kolesnikov, et~al.
\newblock Mlp-mixer: An all-mlp architecture for vision.
\newblock \emph{Advances in Neural Information Processing Systems}, 34, 2021.

\bibitem{melas2021you}
Melas-Kyriazi, L.
\newblock Do you even need attention? a stack of feed-forward layers does
  surprisingly well on imagenet.
\newblock \emph{arXiv preprint arXiv:2105.02723}, 2021.

\bibitem{yu2022metaformer}
Yu, W., M.~Luo, P.~Zhou, et~al.
\newblock Metaformer is actually what you need for vision.
\newblock In \emph{Proceedings of the IEEE/CVF conference on computer vision
  and pattern recognition}, pages 10819--10829. 2022.

\bibitem{yu2022metaformer2}
Yu, W., C.~Si, P.~Zhou, et~al.
\newblock Metaformer baselines for vision.
\newblock \emph{IEEE Transactions on Pattern Analysis and Machine
  Intelligence}, 2023.

\bibitem{hopfield82}
Hopfield, J.~J.
\newblock Neural networks and physical systems with emergent collective
  computational abilities.
\newblock \emph{Proceedings of the National Academy of Sciences},
  79(8):2554--2558, 1982.

\bibitem{hopfield84}
---.
\newblock Neurons with graded response have collective computational properties
  like those of two-state neurons.
\newblock \emph{Proceedings of the National Academy of Sciences},
  81(10):3088--3092, 1984.

\bibitem{krotov2020large}
Krotov, D., J.~J. Hopfield.
\newblock Large associative memory problem in neurobiology and machine
  learning.
\newblock In \emph{International Conference on Learning Representations}. 2021.

\bibitem{ramsauer2021hopfield}
Ramsauer, H., B.~Sch{\"a}fl, J.~Lehner, et~al.
\newblock Hopfield networks is all you need.
\newblock In \emph{International Conference on Learning Representations}. 2021.

\bibitem{tang2021remark}
Tang, F., M.~Kopp.
\newblock A remark on a paper of krotov and hopfield [arxiv: 2008.06996].
\newblock \emph{arXiv preprint arXiv:2105.15034}, 2021.

\bibitem{krotov2021hierarchical}
Krotov, D.
\newblock Hierarchical associative memory.
\newblock \emph{arXiv preprint arXiv:2107.06446}, 2021.

\bibitem{behrmann2019invertible}
Behrmann, J., W.~Grathwohl, R.~T. Chen, et~al.
\newblock Invertible residual networks.
\newblock In \emph{International Conference on Machine Learning}, pages
  573--582. PMLR, 2019.

\bibitem{bai2019deep}
Bai, S., J.~Z. Kolter, V.~Koltun.
\newblock Deep equilibrium models.
\newblock \emph{Advances in Neural Information Processing Systems}, 32, 2019.

\bibitem{ghaoui2021implicit}
El~Ghaoui, L., F.~Gu, B.~Travacca, et~al.
\newblock Implicit deep learning.
\newblock \emph{SIAM Journal on Mathematics of Data Science}, 3(3):930--958,
  2021.

\bibitem{touvron2022resmlp}
Touvron, H., P.~Bojanowski, M.~Caron, et~al.
\newblock Resmlp: Feedforward networks for image classification with
  data-efficient training.
\newblock \emph{IEEE Transactions on Pattern Analysis and Machine
  Intelligence}, 2022.

\bibitem{liu2022we}
Liu, R., Y.~Li, L.~Tao, et~al.
\newblock Are we ready for a new paradigm shift? a survey on visual deep mlp.
\newblock \emph{Patterns}, 3(7):100520, 2022.

\bibitem{rao2021global}
Rao, Y., W.~Zhao, Z.~Zhu, et~al.
\newblock Global filter networks for image classification.
\newblock \emph{Advances in neural information processing systems},
  34:980--993, 2021.

\bibitem{tatsunami2022sequencer}
Tatsunami, Y., M.~Taki.
\newblock Sequencer: Deep lstm for image classification.
\newblock In \emph{Advances in Neural Information Processing Systems}. 2022.

\bibitem{han2022vision}
Han, K., Y.~Wang, J.~Guo, et~al.
\newblock Vision gnn: An image is worth graph of nodes.
\newblock In \emph{Advances in Neural Information Processing Systems}. 2022.

\bibitem{NIPS2016_eaae339c}
Krotov, D., J.~J. Hopfield.
\newblock Dense associative memory for pattern recognition.
\newblock In \emph{Advances in Neural Information Processing Systems}, vol.~29.
  2016.

\bibitem{demircigil2017model}
Demircigil, M., J.~Heusel, M.~L{\"o}we, et~al.
\newblock On a model of associative memory with huge storage capacity.
\newblock \emph{Journal of Statistical Physics}, 168(2):288--299, 2017.

\bibitem{krotov2018dense}
Krotov, D., J.~J. Hopfield.
\newblock Dense associative memory is robust to adversarial inputs.
\newblock \emph{Neural computation}, 30(12):3151--3167, 2018.

\bibitem{NEURIPS2020_da4902cb}
Widrich, M., B.~Sch\"{a}fl, M.~Pavlovi\'{c}, et~al.
\newblock Modern hopfield networks and attention for immune repertoire
  classification.
\newblock In \emph{Advances in Neural Information Processing Systems}, vol.~33,
  pages 18832--18845. 2020.

\bibitem{yang2022transformers}
Yang, Y., Z.~Huang, D.~Wipf.
\newblock Transformers from an optimization perspective.
\newblock In \emph{Advances in Neural Information Processing Systems}. 2022.

\bibitem{hoover2023energy}
Hoover, B., Y.~Liang, B.~Pham, et~al.
\newblock Energy transformer.
\newblock \emph{arXiv preprint arXiv:2302.07253}, 2023.

\bibitem{winston2020monotone}
Winston, E., J.~Z. Kolter.
\newblock Monotone operator equilibrium networks.
\newblock \emph{Advances in neural information processing systems},
  33:10718--10728, 2020.

\bibitem{bai2021stabilizing}
Bai, S., V.~Koltun, Z.~Kolter.
\newblock Stabilizing equilibrium models by jacobian regularization.
\newblock In \emph{International Conference on Machine Learning}, pages
  554--565. PMLR, 2021.

\bibitem{bai2020multiscale}
Bai, S., V.~Koltun, J.~Z. Kolter.
\newblock Multiscale deep equilibrium models.
\newblock \emph{Advances in Neural Information Processing Systems},
  33:5238--5250, 2020.

\bibitem{kawaguchi2021theory}
Kawaguchi, K.
\newblock On the theory of implicit deep learning: Global convergence with
  implicit layers.
\newblock In \emph{International Conference on Learning Representations
  (ICLR)}. 2021.

\bibitem{pmlr-v162-agarwala22a}
Agarwala, A., S.~S. Schoenholz.
\newblock Deep equilibrium networks are sensitive to initialization statistics.
\newblock In \emph{Proceedings of the 39th International Conference on Machine
  Learning}, vol. 162, pages 136--160. 2022.

\bibitem{cifar}
Krizhevsky, A.
\newblock Learning multiple layers of features from tiny images, 2009.

\bibitem{rw2019timm}
Wightman, R.
\newblock Pytorch image models.
\newblock \url{https://github.com/rwightman/pytorch-image-models}, 2019.

\bibitem{loshchilovdecoupled}
Loshchilov, I., F.~Hutter.
\newblock Decoupled weight decay regularization.
\newblock In \emph{International Conference on Learning Representations}. 2019.

\bibitem{Szegedy_2016_CVPR}
Szegedy, C., V.~Vanhoucke, S.~Ioffe, et~al.
\newblock Rethinking the inception architecture for computer vision.
\newblock In \emph{Proceedings of the IEEE Conference on Computer Vision and
  Pattern Recognition (CVPR)}. 2016.

\bibitem{huang2016deep}
Huang, G., Y.~Sun, Z.~Liu, et~al.
\newblock Deep networks with stochastic depth.
\newblock In \emph{Computer Vision--ECCV 2016: 14th European Conference,
  Amsterdam, The Netherlands, October 11--14, 2016, Proceedings, Part IV 14},
  pages 646--661. Springer, 2016.

\bibitem{devries2017improved}
DeVries, T., G.~W. Taylor.
\newblock Improved regularization of convolutional neural networks with cutout.
\newblock \emph{arXiv preprint arXiv:1708.04552}, 2017.

\bibitem{Yun_2019_ICCV}
Yun, S., D.~Han, S.~J. Oh, et~al.
\newblock Cutmix: Regularization strategy to train strong classifiers with
  localizable features.
\newblock In \emph{Proceedings of the IEEE/CVF International Conference on
  Computer Vision (ICCV)}. 2019.

\bibitem{zhangmixup}
Zhang, H., M.~Cisse, Y.~N. Dauphin, et~al.
\newblock mixup: Beyond empirical risk minimization.
\newblock In \emph{International Conference on Learning Representations}. 2018.

\bibitem{zhong2020random}
Zhong, Z., L.~Zheng, G.~Kang, et~al.
\newblock Random erasing data augmentation.
\newblock In \emph{Proceedings of the AAAI conference on artificial
  intelligence}, vol.~34, pages 13001--13008. 2020.

\bibitem{cubuk2020randaugment}
Cubuk, E.~D., B.~Zoph, J.~Shlens, et~al.
\newblock Randaugment: Practical automated data augmentation with a reduced
  search space.
\newblock In \emph{Proceedings of the IEEE/CVF conference on computer vision
  and pattern recognition workshops}, pages 702--703. 2020.

\bibitem{KrauseStarkDengFei-Fei_3DRR2013}
Krause, J., M.~Stark, J.~Deng, et~al.
\newblock 3d object representations for fine-grained categorization.
\newblock In \emph{4th International IEEE Workshop on 3D Representation and
  Recognition (3dRR-13)}. 2013.

\bibitem{bossard14}
Bossard, L., M.~Guillaumin, L.~Van~Gool.
\newblock Food-101 -- mining discriminative components with random forests.
\newblock In \emph{European Conference on Computer Vision}. 2014.

\bibitem{NIPS2012_c399862d}
Krizhevsky, A., I.~Sutskever, G.~E. Hinton.
\newblock Imagenet classification with deep convolutional neural networks.
\newblock In \emph{Advances in Neural Information Processing Systems}, vol.~25.
  2012.

\bibitem{hou2022vision}
Hou, Q., Z.~Jiang, L.~Yuan, et~al.
\newblock Vision permutator: A permutable mlp-like architecture for visual
  recognition.
\newblock \emph{IEEE Transactions on Pattern Analysis and Machine
  Intelligence}, 2022.

\end{thebibliography}
